%% file: main.tex
\documentclass[lettersize,journal]{IEEEtran}
\usepackage{amsmath,amsfonts,mathrsfs,euscript}

\usepackage{array}
\usepackage{subfig}
\usepackage{textcomp}
\usepackage{stfloats}
\usepackage{hyperref}
\usepackage{url}
\usepackage{verbatim}
\usepackage{graphicx}
\usepackage{multirow}
\usepackage[table,xcdraw]{xcolor}
\usepackage{colortbl}
\usepackage{tcolorbox}
\usepackage{lipsum}
\usepackage{pifont}%
\usepackage{tikz}
\usepackage{pgfplots}
\usetikzlibrary{pgfplots.colormaps}
\pgfplotsset{compat=1.8}
\usepackage[utf8]{inputenc}
\usepackage[export]{adjustbox}
\usepackage[ruled,vlined]{algorithm2e}
\usepackage{algpseudocode,amsmath}
\usepackage{textcomp}
\tcbuselibrary{listings,breakable}
\usepgfplotslibrary{colorbrewer}
\usetikzlibrary{pgfplots.statistics, pgfplots.colorbrewer} 
\usepgfplotslibrary{groupplots}

\usetikzlibrary{patterns}
\usepackage{balance}
\newtcolorbox{textbox}{
    colback=white,
    colframe=black,
    boxrule=.5pt,
    breakable,
}

\usepackage{todonotes}
\usepackage[capitalize]{cleveref}
\usepackage{booktabs}
\definecolor{red}{rgb}{1.0, 0.44, 0.37}
\begin{document}
\input{results.dat}

\title{Bio-inspired Agentic Self-healing Framework for Resilient Distributed Computing Continuum Systems}

\author{Alaa Saleh, %~\IEEEmembership{Student Member,~IEEE},
Praveen Kumar Donta, %~\IEEEmembership{Senior Member,~IEEE},
Roberto Morabito, %~\IEEEmembership{Member,~IEEE},
Sasu Tarkoma, %~\IEEEmembership{Senior Member,~IEEE},
Anders Lindgren, %~\IEEEmembership{Senior Member,~IEEE},
Qiyang Zhang,\\ %~\IEEEmembership{Member,~IEEE},
Schahram Dustdar, %~\IEEEmembership{Fellow,~IEEE},
Susanna Pirttikangas, %~\IEEEmembership{Senior Member,~IEEE},
and {L}auri Lov\'en %~\IEEEmembership{Senior Member,~IEEE}

\thanks{A. Saleh (corresponding author), S. Pirttikangas, and L. Lov\'en are with Center for Applied Computing, University of Oulu, Oulu 90014, Finland;}\thanks{P. Donta is with Department of Computer and Systems Sciences, Stockholm University, Stockholm 106 91, Sweden;}\thanks{ R. Morabito is with Department of Communication Systems, EURECOM, Biot 06410, France;}\thanks{ S. Tarkoma is with Department of Computer Science, University of Helsinki, Helsinki 00100, Finland;}\thanks{ A. Lindgren is with RISE Research Institutes of Sweden, Stockholm 166 40, Sweden and Department of Computer Science, Electrical and Space Engineering, Luleå University of Technology, Luleå 971 87, Sweden;}\thanks{Q. Zhang is with School of Computer Science, Peking University, Beijing 100087, China;}\thanks{S. Dustdar is with Distributed Systems Group, TU Wien, Wien 1040, Austria and ICREA, Barcelona, Barcelona 08002, Spain}% 
% \thanks{Manuscript received XX X, 2025; revised XX X, 20XX.}
\thanks{Funding: This work was supported by the Research Council of Finland through the 6G Flagship program (grant 318927) and the CO2CREATION SRC project (grant 372355), by Business Finland through the Neural pub/sub research project (diary number 8754/31/2022), and by the ERDF (project numbers A81568, A91867).
}}

\markboth{ }% 
{Saleh \MakeLowercase{\textit{et al.}}: Bio-inspired Agentic Self-healing Framework for Resilient Distributed Computing Continuum Systems}

\maketitle

\begin{abstract}
Human biological systems sustain life through extraordinary resilience, continually detecting damage, orchestrating targeted responses, and restoring function through self-healing. Inspired by these capabilities, this paper introduces \texttt{\textbf{ReCiSt}}, a bio-inspired agentic self-healing framework designed to achieve resilience in Distributed Computing Continuum Systems (DCCS). Modern DCCS integrate heterogeneous computing resources, ranging from resource-constrained IoT devices to high-performance cloud infrastructures, and their inherent complexity, mobility, and dynamic operating conditions expose them to frequent faults that disrupt service continuity. These challenges underscore the need for scalable, adaptive, and self-regulated resilience strategies. \texttt{\textbf{ReCiSt}} reconstructs the biological phases of Hemostasis, Inflammation, Proliferation, and Remodeling into the computational layers Containment, Diagnosis, Meta-Cognitive, and Knowledge for DCCS. These four layers perform autonomous fault isolation, causal diagnosis, adaptive recovery, and long-term knowledge consolidation through Language Model (LM)-powered agents. These agents interpret heterogeneous logs, infer root causes, refine reasoning pathways, and reconfigure resources with minimal human intervention. The proposed \texttt{\textbf{ReCiSt}} framework is evaluated on public fault datasets using multiple LMs, and no baseline comparison is included due to the scarcity of similar approaches. Nevertheless, our results, evaluated under different LMs, confirm ReCiSt’s self-healing capabilities within tens of seconds with minimum of 10\% of agent CPU usage. Our results also demonstrated depth of analysis to over come uncertainties and amount of micro-agents invoked to achieve resilience.
%{\color{red}The results show improvements in time taken for self-healing by \textbf{X\%}, CPU utilization of agents by \textbf{Y\%}, decision making quality of micro-agents of \textbf{Z\%}, and ABC.}
\end{abstract}


\section{Introduction}
\IEEEPARstart{H}{uman} beings constitute one of the most intelligent biological species on Earth. The human body functioning as a highly optimized distributed ecosystem composed of approximately 37.2 trillion specialized cells. These cells cooperate through organ systems such as the nervous, cardiovascular, respiratory, muscular and skeletal systems, each performing localized computations while contributing to global physiological stability. For example, the nervous system exemplifies distributed processing: the heart contains roughly 40000 neurons that regulate cardiac rhythm, the gut (the second brain) consists of nearly 500 million neurons responsible for autonomous digestive control, each retina contains over 100 million to process visual signals, and the spinal cord hosts millions of neurons that execute low-latency reflexive responses \cite{donta2025human}. This hierarchical yet decentralized biological architecture parallels modern computing paradigm called Distributed Computing Continuum Systems (DCCS) \cite{dustdar2022distributed}, integrates edge devices, intermediate fog nodes and cloud infrastructures into a unified computational fabric capable of allocating tasks based on latency, energy constraints and computational load. DCCS face significant operational challenges due to their heterogeneous and continuously evolving infrastructures. Nodes vary in computational capacity, storage, connectivity, and reliability, which demands orchestration mechanisms capable of adapting to fluctuating workloads, node mobility, and dynamic network conditions. Similar forms of fluctuation are intrinsic to the human body as well.

Considering these similarities, integrating biological models into DCCS offers a promising approach to addressing several of its core challenges, as discussed in \cite{donta2025human}. In this paper, we explore and simulate one such direction by mapping biological self-regulation processes to DCCS to enable self-healing behavior and strengthen system resilience. The human body provides a natural model for distributed self-regulation~\cite{10.1145/1152934.1152937}, as it can detect disruptions, isolate damaged regions, and restore function while maintaining overall stability. Processes such as wound healing, immune response, and distributed neuronal decision making coordinate sensing, feedback, and adaptation across heterogeneous components, demonstrating a scalable and decentralized form of resilience. Adopting these principles into DCCS architectures allows systems to contain failures rapidly, reconfigure affected nodes, and maintain service continuity under dynamic and uncertain conditions. 

\subsection{Biological Basis}
Biological wound healing is a multi-phase process that shows an intrinsic capacity to detect tissue damage, initiate targeted responses, and restore functional integrity without external intervention. To provide necessary biological context for computing-field readers, we outline four phases of wound healing~\cite{enoch2008basic} including Hemostasis, Inflammation, Proliferation, and Remodeling phases. 
\subsubsection{Hemostasis} is the initial phase of wound healing and constitutes the body’s immediate response to vascular injury. After tissue damage, local blood vessels undergo vasoconstriction to limit blood flow and reduce blood loss. Platelets are then activated by thrombin and exposed fibrillar collagen, a process supported by the collagen-associated amino acids proline and hydroxyproline. Activated platelets adhere to the collagen matrix and aggregate to form an initial platelet plug while releasing mediators such as fibrinogen, which promotes further aggregation. Additional mediators enhance adhesion to collagen and recruit more platelets to the injury site. Concurrently, endothelial cells produce prostacyclin to prevent excessive platelet accumulation. The platelet–fibrinogen complex is subsequently converted to fibrin, forming a stabilizing polymeric network. This fibrin mesh creates a hemostatic clot that seals the wound, prevents additional blood loss.
\subsubsection{Inflammation} 
During the inflammatory phase, the body initiates a defense response to prevent infection and clear cellular debris. Vasodilation occurs, increasing blood flow to the wound site and facilitating the delivery of immune cells, oxygen, and essential nutrients. White blood cells migrate to the injured area to eliminate bacteria, pathogens, and damaged cells. Concurrently, various growth factors are released to stimulate tissue repair processes and recruit additional cells involved in healing. 
\subsubsection{Proliferation} phase focuses on healing and reconstruction of the damaged tissue. This phase is characterized by the replacement of the provisional fibrin matrix with a new extracellular matrix composed of collagen fibers, proteoglycans, and fibronectin, thereby reestablishing tissue integrity and functionality. A key event in this stage is angiogenesis, the formation of new capillaries to replace damaged vasculature and ensure adequate oxygen and nutrient supply to regenerating tissue. Fibroblasts, major effector cells of this phase, migrate into the wound site under the influence of factors released by platelets and macrophages. Their migration follows the alignment of fibrillar structures within the extracellular matrix and is facilitated by localized secretion of proteolytic enzymes. Once positioned in the wound, fibroblasts proliferate and synthesize matrix components such as fibronectin, hyaluronan, collagen, and proteoglycans, supporting new matrix construction and cellular ingrowth. Angiogenesis proceeds through oxygen-dependent regulation: hypoxia increases Hypoxia-Inducible Factor (HIF), which induces Vascular Endothelial Growth Factor (VEGF) for neovascularization, whereas reoxygenation degrades HIF and reduces VEGF. As vascularization improves, fibroblast proliferation decreases, epithelialization restores the epidermal barrier, and myofibroblasts contract the wound to reduce its size.
\subsubsection{Remodeling/maturation} 
During the this phase of tissue repair, the newly formed tissue undergoes progressive strengthening, reorganization, and functional refinement. Collagen fibers are realigned and remodeled to enhance the tensile strength and elasticity of the regenerated tissue, contributing to the restoration of structural integrity. Concurrently, the vascular network that had proliferated during earlier stages of healing undergoes regression. As a result, the wound gradually loses its characteristic red or pink coloration, signifying the completion of tissue maturation.

\subsection{Motivation}
\begin{figure}[t]
    \centering
    \includegraphics[width=0.99\linewidth]{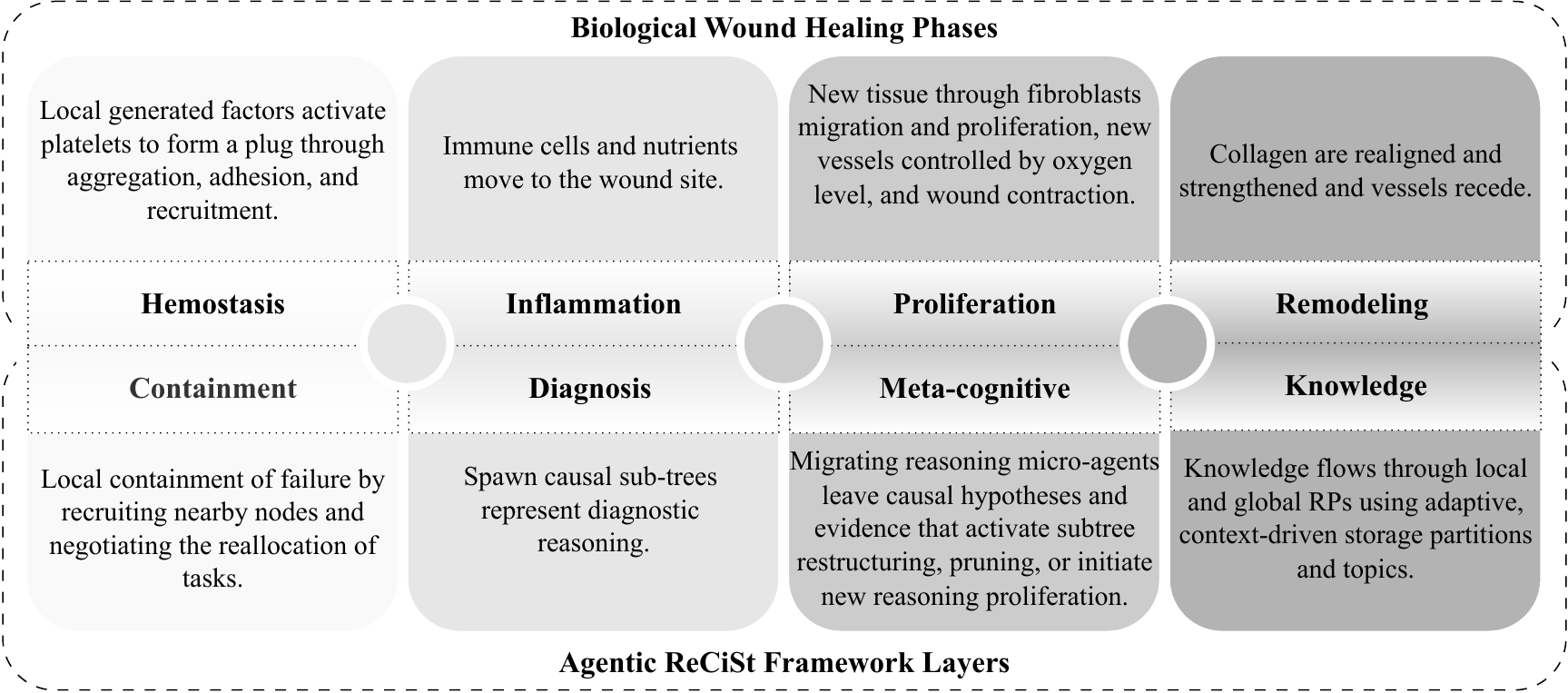}
    \caption{Functional Mapping Biological Wound Healing Phases to Self-healing Layers in the \texttt{ReCiSt} Framework}
    \label{fig:woundhealing}
\end{figure}

Motivated by these observations, we introduce ReCiSt, a bio-inspired agentic self-healing architecture for resilient DCCS. Fig. \ref{fig:woundhealing} illustrates the mapping of biological wound-healing phases to the self-healing layers in the ReCiSt framework. Hemostasis corresponds to the system’s immediate fault response, where isolation and mitigation are initiated; in ReCiSt, this function is performed by the Containment Layer, which negotiates rerouting of affected services to stable neighboring nodes to prevent cascading disruptions. The Inflammation Phase, reflecting the biological immune response, aligns with the Diagnosis Layer, where operational data are collected and analyzed to determine the fault’s nature and scope. In the Proliferation Phase, associated with new tissue and vessel formation, the Meta-Cognitive Layer enables micro-agent proliferation, dynamic reasoning, and the creation of new communication pathways through updated routing tables. Finally, the Remodeling Phase, where biological tissue strengthens, corresponds to the Knowledge Layer, which propagates knowledge across the distributed continuum system through coordinated local and global Rendezvous Points (RP). 

ReCiSt framework is designed to detect disruptions, diagnose their underlying causes, regulate its internal reasoning processes, and optimize its distributed knowledge structures to achieve consistent system performance under uncertain conditions. These capabilities are enabled by Language Models (LMs)-powered agents that execute localized containment, perform causal discovery, regulate internal reasoning through meta-cognitive mechanisms, and manage adaptive knowledge-sharing structures that reorganize in response to contextual drift. Achieving this level of adaptability requires agentic capabilities in each phase to enable systems to regulate their own reasoning processes and adjust internal decision-making structures as operational demands evolve~\cite{10.1145/3774946,10.1145/2168260.2168276}.

\subsection{Contributions}
\begin{itemize}
    \item We propose ReCiSt framework aims to provide an adaptive and  agentic self-healing system that initiates defensive responses through reflexive local containment, and discovers causal dependencies inspired from human body's self healing mechanism, i.e., wound healing.
    \item ReCiSt enables adaptive self-regulation of the agent’s internal reasoning via migratory micro-agents.
    \item ReCiSt supports knowledge sharing through local and global RPs using adaptive, context-driven storage.
    \item Our framework is designed to operate effectively across heterogeneous datasets that vary in scale, structure, failure characteristics, and operational context. 
\end{itemize}
We implement a prototype of the ReCiSt framework and evaluate it on multiple public DCCS datasets using different LMs, showing effective self-healing with reduced recovery time, controlled agent resource overhead, and improved decision quality.

\section{Related works} \label{sec:Literature}
Resilience is becoming a key research focus in distributed infrastructures, especially within the communications community. Recent 6G roadmaps explicitly prioritize resilience as a core standardization target for future network architectures \cite{alves20256gresiliencewhite}. Altaweel et al. \cite{altaweel2022rsock} propose an identity-based routing protocol for mission-critical fog and edge deployments in adaptive routing under dynamic network conditions. Similarly, Nakayama et al. \cite{nakayama2022resilience} develop a resilient architecture for multipath communication in mobile networks. In \cite{10.1145/3725985}, a log-based fault tolerance for dynamic workloads has been explored through serverless runtime designs.

Notable contributions foreground resilience through data- and workload-oriented strategies that exploit machine learning (ML) for prediction and coordination. Sen et al. \cite{sen2021resilient} develop a resilient edge–cloud architecture that combines server failure prediction with optimized virtual machine (VM) migration and multi-hop routing to ensure seamless service continuity. Díaz et al. \cite{diaz2023akats} employ federated ML failure prediction together with optimization heuristics to select deployment configurations that improve fault tolerance in edge workloads, while Kashyap et al. \cite{kashyap2025predictive} focus on proactive resource allocation in dynamic fog environments by forecasting per-task resource demands and guiding task partitioning to mitigate node failures. Through these solutions are fault-tolerance, but extensive training data and learning cycles, limiting their responsiveness to immediate or unpredictable system changes. 

Some other recent works underscores the difficulty of distinguishing faults from slowdowns under uncertain conditions. For example, \cite{9466159} proposes gossip-based communication with migrating agents as an effective decentralized detection mechanism. Subsequent frameworks introduce knowledge-driven and dynamically generated self-healing agents capable of prediction, diagnosis, and autonomous service redeployment \cite{10562327}. Additional advances automate the generation and evolutionary optimization of multi-agent workflows \cite{wang2025ev} and develop structured context-management architecture that equip Large Language Model (LLM)-driven agents with memory as version-controlled file system for coherent distributed reasoning \cite{wu2025git}. Emerging self-learning paradigms further couple task generation, policy optimization, and reward evaluation into closed-loop processes that iteratively refine agent capabilities \cite{sun2025agent}. LLM-driven multi-agent systems with specialized roles extend these capabilities by jointly analyzing traffic patterns, monitoring performance, and detecting suspicious activities to determine optimized mitigation strategies for adaptive network management \cite{11169757}.

In communication networks, agentic mechanisms employ intent-aware reasoning to support real-time resource allocation under fluctuating conditions \cite{Wu_2025}. At the same time, diagnosis pipelines that integrate hierarchical reasoning, multi-pipeline, and fine-tuned smaller LLMs improve root-cause analysis and failure localization in cloud networks \cite{10.1145/3718958.3750505}, while LLM-based, proactive fault-tolerance frameworks in edge networks orchestrate containment, diagnosis, and recovery tools to mitigate failures before services degrade \cite{10620727}. These research attempt to achieve fully autonomous and embedding self-improved agent mechanisms, yet not achieved desired solutions within resource limit environments.  

Given our emphasis on bio-inspired strategies, we observed that existing research in this area remains relatively sparse. Fro example, \cite{10.1145/3575879.3575976} demonstrates how biological mechanisms can help continuous network reorganization to maintain scalability and robustness in heterogeneous sensing environments. Similar inspiration in autonomic network-management by mapping molecular biology onto system- and device-level control processes \cite{10.1145/1315843.1315880}, and enable hardware to evolve and repair itself in response to structural faults \cite{10.1145/1570256.1570291}. Multi-agent, bio-inspired framework further shows how self-managing entities negotiate to restore performance during complex scheduling disruptions \cite{10.1145/1388969.1389045}. Recent communication-network architectures integrate evolutionary, immune, and neural models to accelerate fault detection and repair \cite{11064129}, and biologically inspired machine-learning frameworks apply swarm and immune metaphors to achieve rapid, distributed recovery from failures \cite{11196711}. Active inference has been proposed as a resilience strategy for DCCS~\cite{donta2025resilientdesignactive}, rooted in bio-inspired models that emulate the human brain’s continuous reasoning and adaptation.

Most existing literature focus on efficient fault detection and autonomous recovery supported by AI- or ML-based decision mechanisms. However, these approaches are often constrained by their dependence on pretrained models, externally supplied decision policies, and large datasets, which limits their ability to handle previously unseen or evolving faults. Furthermore, their computational and data-intensive nature reduces practicality in distributed environments with heterogeneous resource capacities. Overcoming these limitations requires self-healing mechanisms that can reason about disruptions, adapt internal decision processes, and dynamically reorganize knowledge structures to sustain performance under uncertain  conditions.

\section{System Model and Problem formulation}\label{sec:systemodel}
A {DCCS} can be formally represented as a graph $G = (\mathcal{N}, E)$, where $\mathcal{N} = \{N_1, N_2, \ldots, N_n \}$ is the set of heterogeneous nodes (spanning IoT devices, edge, fog, and cloud resources) and $E$ is the set of communication links between them. Each node $N_i$ is described by its attributes $(\mathcal{C}_i, \mathcal{M}_i, \mathcal{S}_i, \mathcal{V}_i)$, where $\mathcal{C}_i$ denotes computational capacity (such as CPU/GPU cycles), $\mathcal{M}_i$ is the device storage status, the device condition such as down, available (space available to run more tasks), busy (resource occupied by tasks), and recovering represents $\mathcal{S}_i\in \{00,11,01,10\}$, and $\mathcal{V}_i\in \{00,11,01,10\}$ indicates corresponding to \{low, medium, high, critical\} vulnerability. Tasks $\mathcal{T} = \{\tau_1, \tau_2, \ldots, \tau_m \}$ arrive dynamically and need to be mapped to $N_i$ for execution. Assume that a node $N_i$ can compute one or more $\mathcal{T}$ depends on its $(\mathcal{C}_i, \mathcal{M}_i, \mathcal{S}_i, \mathcal{V}_i)$. The task allocation at time $t$ can be specified by an allocation function $A(t): \mathcal{T} \rightarrow \mathcal{N},\ A(t, \tau_j) = N_i $, such that $A(t, \tau_j) = N_i$ assigns task $\tau_j$ to node $N_i$ if $\mathcal{S}_i = \{10\}$ along with $c_j\leq C_i$ and $m_j \leq M_i$. Each communication link between node $N_i$ and $N_j$ is $ e_{ij} \in E $ has a bandwidth $ b_{ij} \geq \delta$, where $\delta$ is bandwidth threshold. The end-to-end latency of task $\tau_j$ under allocation function $\mathcal{A}$ is defined as $l_j (A) \leftarrow l_j^{net} (A) + l_j^{com}(A)$, where $l_j^{net} (A)$ is network latency and $l_j^{com}(A)$ is computational latency. The overall latency of the system is $\mathcal{L}(A) = \frac{1}{m}\sum_{j=1}^{m} l_j(A)$.

Failures in the DCCS are captured through failure scenarios $\omega$, where $F_\omega(t)\subseteq\mathcal{N}$ denotes the set of nodes that become unavailable at time $t$. When a node $N_i \in F_\omega(t)$, its operational state transitions to $\mathcal{S}_i(t)=\text{down}$, which disrupts all tasks currently assigned to it (i.e., those satisfying $A(t,\tau_j)=N_i$). Once the self-healing process is initiated, the node enters a recovering state,   $\mathcal{S}_i(t^+)=\text{recovering}$, and the system executes a healing procedure that detects the failure, isolates the affected tasks, and reallocates them to surviving nodes whose operational states and resources allow execution. This produces an updated allocation defined as     $A_\omega(t^+) = \text{Heal}\!\left(A(t),F_\omega(t)\right)$.
For each task $\tau_j$ under scenario $\omega$, we define the completion indicator $I_j(A,\omega)=\{0~or~1\}$, where 1 if $\tau_j$ successfully completes under $A_\omega(t^+)$, 0 otherwise, and compute the resilience of the allocation strategy as
\begin{equation}
    \label{eq:resilience}
    \mathcal{R}(A)=
    \mathbb{E}_\omega \left[
        \frac{1}{m}\sum_{j=1}^{m} I_j(A,\omega)
    \right],
\end{equation}
which quantifies the expected fraction of tasks that the system can successfully complete despite node failures, state transitions (down $\rightarrow$ recovering), and the associated self-healing operations. Resource utilization is defined by jointly considering CPU and memory usage as
\begin{equation}
    \mathcal{U}(A) =
    \alpha
    \frac{\sum_{i\in\mathcal{N}} \text{cpu\_load}(N_i)}{\sum_{i\in\mathcal{N}} \mathcal{C}_i}
    +
    (1-\alpha)
    \frac{\sum_{i\in\mathcal{N}} \text{mem\_load}(N_i)}{\sum_{i\in\mathcal{N}} \mathcal{M}_i},
    \label{eq:utilization}
\end{equation}
where $0 < \alpha \leq 1$ controls the relative importance of compute and memory utilization.
The DCCS controller aims to determine an allocation strategy that jointly optimizes latency, resource utilization, and resilience. The resulting multi-objective optimization problem is formulated as
\begin{subequations}
\begin{align}
    &\min_{A}\ \mathcal{L}(A), 
    \qquad
    \max_{A}\ \mathcal{U}(A), 
    \qquad
    \max_{A}\ \mathcal{R}(A),
    \label{eq:objective}
\end{align}\vspace{-0.6cm}
\begin{flalign*}
\text{subject to:} &&
\end{flalign*}\vspace{-0.8cm}
\begin{align}
    &\sum_{\tau_j \in \mathcal{T}: A(t,\tau_j)=N_i} c_j 
      \le \mathcal{C}_i,
    \qquad \forall N_i \in \mathcal{N},
    \label{eq:objective:a} \\
    &\sum_{\tau_j \in \mathcal{T}: A(t,\tau_j)=N_i} m_j 
      \le \mathcal{M}_i,
    \qquad \forall N_i \in \mathcal{N},
    \label{eq:objective:b} \\
    &\mathcal{S}_{A(t,\tau_j)} = \text{available},
    \qquad \forall \tau_j \in \mathcal{T},
    \label{eq:objective:c} \\
    &\mathcal{V}_{A(t,\tau_j)} \in \{\text{low}, \text{medium}\}
    \quad \text{for critical tasks } \tau_j,
    \label{eq:objective:d} \\
    &b_{ij} \ge \delta 
        \qquad \forall e_{ij} \in E,
    \label{eq:objective:e} \\
    &A(t,\tau_j) \in \mathcal{N},
    \qquad \forall \tau_j \in \mathcal{T}.
    \label{eq:objective:f}
\end{align}
\end{subequations}

\section{The ReCiSt Framework}
\begin{figure*}[t]
\centering
\includegraphics[width=0.99\textwidth,keepaspectratio]{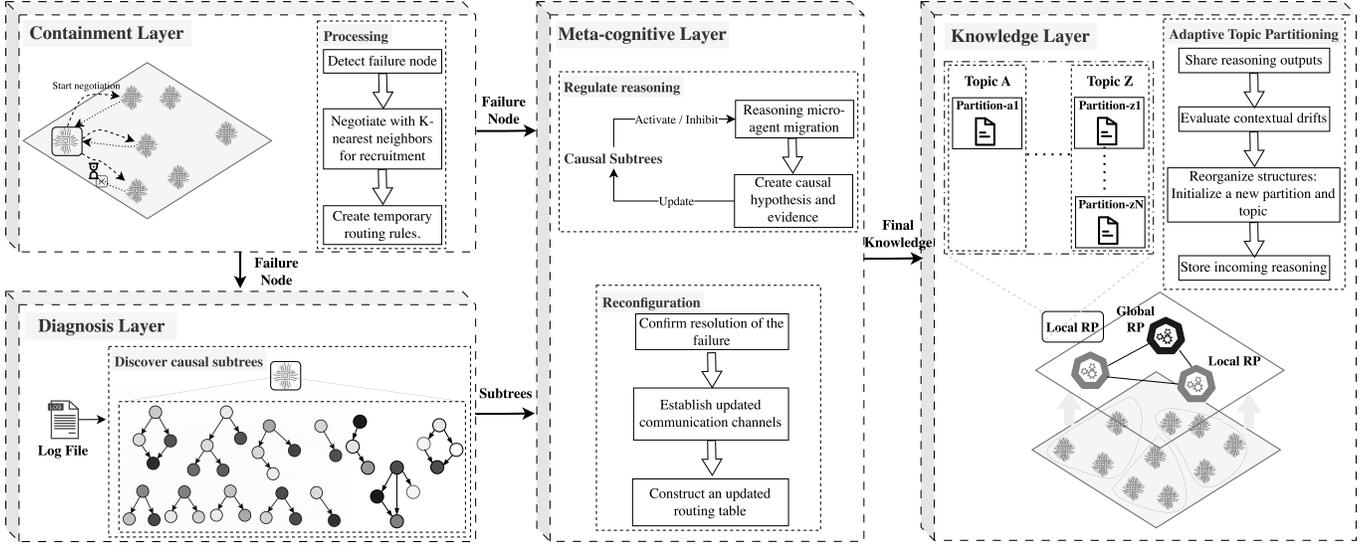}
\caption{A detailed ReCiSt framework illustrating different agentic layers (containment layer, diagnosis layer, meta-cognitive layer, knowledge layer)}\label{fig:frame}
\end{figure*}

As described in the motivation, the ReCiSt framework operates through a self-healing pipeline comprising the Containment, Diagnosis, Meta-cognitive, and Knowledge layers, shown in Fig.~\ref{fig:frame}. 

\subsection{Containment Layer}
This Layer functions as the initial defense and immediate response to make sure uninterrupted services in the ReCiSt framework (top-left of Fig.~\ref{fig:frame} and Algorithm.\ref{Layer1}). 
The system must accurately determine when system deviates from its healthy operational behavior. Due to the heterogeneity of devices, their diverse logging formats \cite{donta2023governance}, and inconsistent sensing capabilities across the continuum, ReCiSt uses LM-driven monitoring agents $\alpha=\{\alpha_1,\alpha_2,\ldots\}$ to perform continuous system state monitoring. Each agent $\alpha_\imath$ maintains updated information regarding the operational state $\mathcal{S}_i$, vulnerability level $\mathcal{V}_i$, and active task set $\mathcal{T}_i$ of all nodes within its $k$-neighborhood $\mathcal{N}_\imath^{(k)}$. Fault identification begins when $\alpha_\imath$ periodically broadcasts a lightweight control signal to every node $N_i \in \mathcal{N}_\imath^{(k)}$. Each node must acknowledge this probe within a predefined interval $\Delta t$ by reporting its instantaneous state $\mathcal{S}_i(t)$ and a minimal heartbeat vector summarizing its current operational load. If a node fails to respond within $\Delta t$, meaning no acknowledgment is received from $N_i$ during the probe window, agent $\alpha_\imath$ classifies this as an abnormal deviation from expected behavior and triggers. It is important to note that at this stage the system does not yet know the exact cause or type of failure. The detected deviation merely indicates that the node has departed from its nominal operating conditions. All such nodes are preliminarily marked as faulty by the Containment Layer, it is included in a failure set denoted by $    F(t)=\{\,N_i \mid \text{Containment Layer flagged } N_i \text{ at time } t\,\}.$
\begin{algorithm}[!t]
\caption{Containment Layer of ReCiSt Framework}\label{Layer1}

\KwIn{$N_i$ \Comment{$N_i$ is number of nodes}} 
\KwOut{$F(t)$ \Comment{failure set}}
\BlankLine
$p = Null$ \Comment{$p$ accepted nodes, initially empty}
\begin{tcolorbox}[colback=white!100!black, colframe=white!90!black, coltitle=black, title= Agent Prompt, width=0.45\textwidth]
- CALL TOOL $\rightarrow$ CHECK status of neighbors $\mathcal{N}_\imath^{(k)}$:\\
\ForEach{$N_i \in \mathcal{N}_\imath^{(k)}$}{ 
   \If{$N_i(t) \geq \Delta t$}{
      $\mathcal{S}_i(t)=00$\\
      $F(t) \leftarrow F(t) \cup N_i$
      }
    } 
- DECIDE non-fault neighborhood set $k_N$:\\
$k_N=\mathcal{N}_\imath^{(k)} - F(t)$ \Comment{$k_N$ is k-nearest neighbors} \\
- CALL Neighbor Agent $\rightarrow$ Negotiation\\
\ForEach{$N_j \in k_N$}{
\begin{tcolorbox}[colback=white!100!black, colframe=white!90!black, coltitle=black, title= Neighbor Agent Prompt, width=0.95\textwidth]
DECIDE\_AND\_RETURN 11, 01, or 10 $\mathcal{S}_j(t) \leftarrow \mathcal{C}_j \& \mathcal{M}_j$ \Comment{Computational capacity and memory of $N_j$}
\end{tcolorbox}
}
- PARSE the response:\\ \If{$\mathcal{S}_j(t) == 11$}{
        $p \leftarrow p \cup {N_j}$ 
    }
- ANALYZE $N_i$ tasks and dependencies. \\
- DECIDE assigning $N_i$ tasks to $p$ 
\end{tcolorbox}

RETURN $F(t)$

\end{algorithm}

When a node $N_i$ is detected when its state becomes $\mathcal{S}_i(t)=00$ for any value of $\mathcal{V}_i$, which triggers the node’s \textit{internal Neural Reflex System} to broadcast a localized containment request to its $k_N$ nearest candidate nodes, represented as the set $k_N=\{N_j \mid N_j \text{ receives a reflex request from agent} \}$. Each node $N_j \in k_N$ responds by reporting its $\mathcal{C}_j$ and $\mathcal{M}_j$, along with  $\mathcal{S}_j(t)=\{11\}$. The Containment Layer request these responses to identify neighbors suitable for redistributing the task set $\mathcal{T}_i$ that was previously running on $N_i$. 
A suitability condition is satisfied when all responding neighbors report $\mathcal{S}_j(t)=11$ for every $N_j$, after which node $N_i$ transitions to an available coordination state $\mathcal{S}_i(t^+)=10$. If the computational capacity and memory of any single neighbor, denoted $\mathcal{C}_j$ and $\mathcal{M}_j$, are insufficient for executing the full task set $\mathcal{T}_i$, meaning $\mathcal{C}_j < \mathcal{C}_i$ of $\mathcal{T}_i$ or $\tau_x\forall x\leq |\mathcal{T}_i|$ and $\mathcal{M}_j < \mathcal{M}_i$ of $ \mathcal{T}_i$ or $\tau_y\forall y\leq |\mathcal{T}_i|$, then the Containment Layer identifies a subset of $p$ nodes from the $k_N$ candidates to cooperatively distribute and execute the workload of $\mathcal{T}_i$ or $\tau_x$.
Through this process, they cooperate to establish a dynamic plug structure that enables the formation of temporary routing rules. The main agent analyzes task dependencies and redistributes the $N_i$ computational workload across the $N_j$ to reduce the risk of cascading failures. The identified failed nodes are transmitted to subsequent layers for further analysis and cure.

% Algorithm.\ref{Layer1} illustrates the workflow of the containment layer within the ReCiSt framework. The agent identifies neighbor nodes in the network that have become unresponsive, classifying any node that fails to return a timely acknowledgment as a failure node $fn$. It then removes these failed nodes from the k-nearest-neighbor set, producing the final candidate collaborator set $K$.  Each candidate node is then queried for its workload status to determine whether it can accept additional tasks. Nodes indicating sufficient capacity are collected as eligible recipients $AcceptedNodes$. Once this subset is established, the agent evaluates the tasks of failure node and dependency structures associated with the failed nodes to determine an appropriate redistribution strategy.

\subsection{Diagnosis Layer}
Once the Containment Layer signals an abnormal deviation, the system transitions into the Diagnosis Layer (bottom-left of Fig.~\ref{fig:frame}), which corresponds to the inflammation stage in biological wound healing. In human physiology, inflammation functions as the initial analytical response, wherein leukocytes migrate to the wound site, identify the nature of the damage, and classify its severity before any tissue repair begins. Analogously, in the ReCiSt framework, the Diagnosis Layer determines the underlying cause, scope, and structural characteristics of the deviated nodes identified in the Containment Layer. At this stage, the system performs structured causal examination to produce a precise, machine-interpretable representation of the failure.

\begin{algorithm}[!t]
\caption{Diagnosis Layer of ReCiSt Framework}\label{Layer2}

\KwIn{$F(t)$}

\ForEach{$N_i~\in~F(t)$}{
/Agent utilize $\mathcal{L}_i$ of $N_i$/ \\
 
\begin{tcolorbox}[colback=white!100!black, colframe=white!90!black, coltitle=black, title= Agent Prompt, width=0.45\textwidth]
$\mathcal{X}_i \leftarrow Null$   \Comment{Node extraction} \\
\ForEach{$l \in \mathcal{L}_i$}{
      EXTRACT $x_{ij}$ \Comment{$x_{ij}$ is events, metrics, states, components}\\
      ADD each $x_{ij} to \mathcal{X}_i$ 
      }
$\mathcal{E}_i \leftarrow Null$  \Comment{Causal decomposition} \\
\ForEach{$(x_{ia},x_{ib}) \in \mathcal{X}_i$}{
\If{$\Phi(x_{ia},x_{ib})=1$}{
          $\mathcal{E}_i$.ADD($x_{ia} \rightarrow x_{ib}$)
          }
          }
$G_i^{diag} \leftarrow Null$ \Comment{Return structured subtrees}\\
$G_i^{diag} = (\mathcal{X}_i,\mathcal{E}_i)$ \\
CALL Algorithm 3
\end{tcolorbox}
}
\end{algorithm}

For each $N_i\in F(t)$, the Diagnosis Layer retrieves the corresponding system, network, custom logs, etc., generated during the interval $[t-\Delta_d,t]$, where $\Delta_d$ is the diagnosis window size. Let $\mathcal{L}_i$ denote the aggregated log structure for node $N_i$: $\{\mathcal{L}_i=\mathcal{L}_i^{sys}\cup \mathcal{L}_i^{net}\cup \mathcal{L}_i^{cust} \cup ... \}$
with each component capturing distinct operational dimensions.

The Diagnosis Layer transforms raw log entries into a structured set of observable entities. Each entity corresponds to an event, performance metric, internal state transition, or resource indicator associated with the failure. These entities are instantiated as nodes within a diagnosis variable set $\mathcal{X}_i=\{x_{i1},x_{i2},\ldots,x_{im}\},$ 
which forms the foundation of a graph-based representation of the fault. Each $x_{ij}\in\mathcal{X}_i$ is a symbolic or numeric descriptor derived from $\mathcal{L}_i$, such as CPU anomalies, memory spikes, link degradation indicators, task stalls, or error codes.
Next, this layer infer causal dependencies among elements of $\mathcal{X}_i$. These dependencies define how one internal change leads to another, forming a causal structure that characterizes the faults or abnormal operations. For every ordered pair $(x_{ia},x_{ib}) \forall a, b \in m$, the system evaluates whether a causal relationship exists. A causal relation is encoded as:
\begin{equation}
    x_{ia}\rightarrow x_{ib}\quad \text{iff}\quad \Phi(x_{ia},x_{ib})=1,
\end{equation}
where $\Phi(\cdot)$ is an LM-driven relation-identification function that integrates temporal precedence, log semantics, and learned causal priors.
All these causal relations identified for $N_i$ form a directed edge set:
\begin{equation}
    \mathcal{E}_i=\{\, (x_{ia},x_{ib}) \mid \Phi(x_{ia},x_{ib})=1 \,\}.
\end{equation}
The pair $(\mathcal{X}_i,\mathcal{E}_i)$ therefore constructs a directed graph $G_i^{diag}$:
\begin{equation}\label{eq:causalGraph}
    G_i^{diag} = (\mathcal{X}_i,\mathcal{E}_i),
\end{equation}
which represents the fine-grained causal structure underlying the observed malfunction.

To enhance robustness, the Diagnosis Layer employs an ensemble of parallel reasoning sub-trees. Each sub-tree corresponds to a specialized causal reasoning pathway trained to capture a specific dependency type, such as resource overload, network instability, task-level contention, thermal anomalies, or firmware events. Let the ensemble be represented by $\Psi_i=\{\psi_i^{(1)},\psi_i^{(2)},\ldots,\psi_i^{(q)}\}$,
where $\psi_i^{(k)}$ is the $k$-th sub-tree applied to node $N_i$. Each $\psi_i^{(k)}$ extracts a localized causal subgraph $T_i^{(k)}$ from $G_i^{diag}$:
\begin{equation}
    T_i^{(k)}=(\mathcal{X}_i^{(k)},\mathcal{E}_i^{(k)}), \quad \text{with}\ \mathcal{X}_i^{(k)}\subseteq\mathcal{X}_i,\ \mathcal{E}_i^{(k)}\subseteq\mathcal{E}_i.
\end{equation}
The union of all such sub-trees forms the consolidated diagnosis structure,
\begin{equation}
    \bar{G}_i^{diag}=\bigcup_{k=1}^{q}T_i^{(k)},
\end{equation}
which encodes multilevel, multi-causal interpretations of the failure. This consolidated graph is stored in the node’s diagnosis memory and later transmitted to the Meta-cognitive Layer for reasoning-path restructuring and micro-agent generation. By now, each fault node $N_i$ or link is represented not merely as an isolated malfunction but as a structured causal system that explains \textit{why} the perceived deviation occurred. %This structured representation ensures that subsequent layers can examine, reorganize, and heal the system with higher precision and significantly reduced ambiguity.

\subsection{Meta-Cognitive Layer}
Following the diagnostic phase, the system enters the Meta-Cognitive Layer (middle of Fig.~\ref{fig:frame}), which corresponds to the proliferation stage in biological wound healing. In human tissue, proliferation involves fibroblast activity, extracellular matrix deposition, and angiogenesis that collectively restore structural support and reestablish functional connectivity around the wound site. In the ReCiSt framework, the Meta-Cognitive Layer plays an analogous role at the cognitive level: it reorganizes and extends the causal structures derived from the Diagnostic Layer, manages the generation and regulation of reasoning micro-agents, and refines explanatory pathways so that the system can progress from a raw fault description to a set of coherent and operationally useful hypotheses. The procedural flow of this layer is depicted in Algorithm~\ref{Layer3}.

\begin{algorithm}[!t]
\caption{Meta-cognitive Layer of ReCiSt Framework}\label{Layer3}
\KwIn{$G_i^{diag},N_i$}
\KwOut{$Updatedknowledge$ \Comment{(failure's topic, reason and solution)}}
- CALL TOOL $\rightarrow$ Depth-first traversal:\\
$P^{(k)} \leftarrow$ Depth-first traversal($G_i^{diag}$) \Comment{Create all possible paths from sub-trees}

\ForEach{$p^{(k)} \in P^{(k)}$ }{
\begin{tcolorbox}[colback=white!100!black, colframe=white!90!black, coltitle=black, title= Micro-agent Prompt, width=0.45\textwidth] 
- CALL micro-agents $\mathcal{A}_i^{micro} \rightarrow$ Generate causal candidates $a_i^{(1)},a_i^{(2)},\ldots,a_i^{(r)}$\\
$H_i^{(k)}$ $\leftarrow$ $a_i^{(k)}$.GENERATE($H_i^{(k)} = \Lambda(p^{(k)},\mathcal{L}_i),$) 

\end{tcolorbox}
\begin{tcolorbox}[colback=white!100!black, colframe=white!90!black, coltitle=black, title= Evaluator-agent Prompt, width=0.45\textwidth] 
score $\Gamma(H_i^{(k)})$  $\leftarrow$ PARSE($w_1 C_{coh}(H_i^{(k)}) + w_2 C_{safe}(H_i^{(k)}) + w_3 C_{util}(H_i^{(k)}),$)\\
PARSE\_AND\_DECIDE($\Gamma(H_i^{(k)})$):\\
\If{$\Gamma(H_i^{(k)}) < \theta_{pro}$ \Comment{score is harmful}}{
                CALL $\mathcal{A}_i^{micro}$
         }
\If{score is accept}{
         STORE\_CANDIDATE(status="supporting")\\
         CALL micro-agent($p+1$)
         }
\If{score is reject}{
         CALL micro-agent($p+1$)
         }
\If{$\Gamma(H_i^{(k)}) \geq \theta_{inh}$ \Comment{score is best}}{
          $\mathcal{K}_i^{meta} \leftarrow$ $\mathcal{K}_i^{meta} \cup$ $H_i^{*} = \arg\max_{k} \Gamma(H_i^{(k)})$\\
         $F(t) \leftarrow F(t)~\cap~N_i$
         }
\end{tcolorbox}        
}
RETURN $\mathcal{K}_i^{meta}$
\end{algorithm}

% {\color{blue}Depth-first traversal: It uses a stack that stores the current node, the taken path, and the set of visited nodes. At each iteration, it removes the top stack entry and either records the path if a depth limit has been reached or if the node has no outgoing edges. Otherwise, it examines all outgoing neighbors and adds new stack entries for those not yet visited, extending the path and updating the visited set to avoid cycles. This continues until the stack is empty, yielding a collection of all simple paths starting from the initial node.}

This Layer refine Eq.(\ref{eq:causalGraph}) into a set of viable explanatory and corrective pathways. To accomplish this, the main-agent associated with $N_i$ spawns a population of reasoning micro-agents as shown in Eq.~\ref{eq:microAgent},
\begin{equation}\label{eq:microAgent}
    \mathcal{A}_i^{micro}=\{a_i^{(1)},a_i^{(2)},\ldots,a_i^{(r)}\},
\end{equation}
where $r$ is determined adaptively based on the complexity of the causal graph.
% system resource constraints.
Each micro-agent $a_i^{(k)}$ traverses a path using Depth-first search (DFS) within $\bar{G}_i^{diag}$, seeking to construct a causal explanation of the fault. A path is defined as an ordered sequence of diagnostic variables:
\begin{equation}
    p^{(k)} = (x_{i\ell_1},x_{i\ell_2},\ldots,x_{i\ell_\eta}), \quad \text{with } (x_{i\ell_j},x_{i\ell_{j+1}})\in\mathcal{E}_i,
\end{equation}
where $\eta$ is the depth explored by the micro-agent. For each path $p^{(k)}$, the micro-agent produces a structured hypothesis 
\begin{equation}
    H_i^{(k)} = \Lambda(p^{(k)},\mathcal{L}_i),
\end{equation}
where $\Lambda(\cdot)$ is an LM-driven operator that integrates the path structure with supporting evidence extracted from the logs $\mathcal{L}_i$. Each hypothesis $H_i^{(k)}$ is assigned a meta-cognitive evaluation score computed through Eq.~\ref{eq:Score}
\begin{equation}\label{eq:Score}
    \Gamma(H_i^{(k)}) = w_1 C_{coh}(H_i^{(k)}) + w_2 C_{safe}(H_i^{(k)}) + w_3 C_{util}(H_i^{(k)}),
\end{equation}
where $C_{coh}$ measures causal coherence, $C_{safe}$ quantifies safety of the inferred solution, $C_{util}$ reflects operational feasibility, and $(w_1,w_2,w_3)$ are normalization weights.

The core adaptive mechanism of the Meta-Cognitive Layer is the regulation of micro-agent proliferation, guided by the feedback produced by $\Gamma(H_i^{(k)})$. Analogous to fibroblasts proliferating more rapidly when matrix integrity is low, the system increases the population of micro-agents when current hypotheses exhibit low confidence or poor safety. Formally, proliferation is triggered when $\Gamma(H_i^{(k)}) < \theta_{pro}$, where $\theta_{pro}$ is the proliferation threshold. When triggered, the set $\mathcal{A}_i^{micro}$ is expanded, and new exploratory paths are generated by inserting auxiliary nodes or diverting traversal directions in $\bar{G}_i^{diag}$. This parallels biological angiogenesis, where new vessels extend into damaged regions to restore connectivity; here, new inferential edges are added to enhance the diversity and depth of cognitive exploration.

Conversely, when hypotheses exhibit high confidence and safety, the system inhibits further proliferation by enforcing the condition $\Gamma(H_i^{(k)}) \geq \theta_{inh}$, which suppresses additional micro-agent generation. This feedback-driven balance between activation and inhibition stabilizes the reasoning ecosystem and prevents unnecessary expansion of computational effort.
As micro-agents accumulate evidence and refine hypotheses, the causal graph $\bar{G}_i^{diag}$ undergoes structural reorganization. Let $\Delta\mathcal{X}_i^{(k)}$ and $\Delta\mathcal{E}_i^{(k)}$ denote modifications induced by hypothesis $H_i^{(k)}$. The updated cognitive structure becomes Eq.~\ref{eq:NewCogStructure},
\begin{equation}\label{eq:NewCogStructure}
    \bar{G}_i^{meta} = \left(\mathcal{X}_i \cup \bigcup_{k}\Delta\mathcal{X}_i^{(k)},\, \mathcal{E}_i \cup \bigcup_{k}\Delta\mathcal{E}_i^{(k)}\right).
\end{equation}
Eq.~\ref{eq:NewCogStructure} is directly analogous to the biological formation of new provisional tissue that strengthens the wound region.
Once $\bar{G}_i^{meta}$ reaches a stable configuration, the Meta-Cognitive Layer identifies an optimal hypothesis:
\begin{equation}
    H_i^{*} = \arg\max_{k} \Gamma(H_i^{(k)}),
\end{equation}
which serves as the definitive causal explanation and preliminary corrective strategy for node $N_i$. This output is then propagated to the Knowledge Layer, which corresponds to the remodeling stage in wound healing and is responsible for global system realignment, routing updates, and reintegration of the recovered node.

\subsection{Knowledge Layer}
After the Meta-Cognitive Layer selects an optimal hypothesis $H_i^{*}$ for a failed node $N_i$, the system enters the Knowledge Layer (right of Fig.~\ref{fig:frame}), which mirrors the remodeling phase in biological wound healing. In physiology, remodeling strengthens the extracellular matrix, reorganizes tissue fibers, and integrates the repaired region back into the larger functional structure. Analogously, the Knowledge Layer consolidates, restructures, and disseminates the refined causal and corrective knowledge produced during recovery, thereby supporting stable long-term adaptation within the DCCS. The operational workflow of this layer is detailed in Algorithm~\ref{Layer4}.

\begin{algorithm}[t]
\caption{Knowledge Layer of ReCiSt Framework}\label{Layer4}

\KwIn{$\mathcal{K}_i^{meta}$}
\KwOut{$\mathcal{Z}^{RP} = \{Z_1, Z_2, Z_j,..., Z_u\}$ \Comment{Knowledge stored in memory}}

$\mathbf{e}_{topic} = \phi_{topic}(z_i) ~\forall ~\mathcal{K}_i^{meta} $ \Comment{Compute embeddings for new topic using "text-embedding-3-small" model}\\
$M_T(z_i, Z_j) =\text{STS}\bigl(\mathbf{e}_{topic}, \mathbf{e}_{Z_j}\bigr)$ \Comment{Compute similarity through STS}\\

\If{$\max_{j} M_T(z_i, Z_j) < \theta_{topic}$}{
        $\mathcal{Z}^{RP} \leftarrow \mathcal{Z}^{RP} \cup \{Z_{u+1}= \{P_{u+1}^{(1)}\}$\\
        \Else{
             $\mathbf{e}_{reason} = \phi_{reason}(H_i^{*})$
             $M_R = \text{STS}\bigl(\mathbf{e}_{reason}, \mathbf{e}_{P_{j^*}^{(k)}}\bigr)$\\
             \If{$\max_{k} M_R < \theta_{reason}$}{
                $Z_{j^*}$ $\cup P_{j^*}^{(m_{j^*}+1)}$
             }
             }
             }

RETURN $\mathcal{Z}^{RP}$

\end{algorithm}
The Knowledge Layer is organized around a collection of local and global RPs, which function as adaptive coordination and storage nodes for distributed agents. Each RP maintains a structured knowledge base composed of topic-oriented segments. A topic corresponds to a failure class, and its associated representations include the causal explanations, hypotheses, and corrective strategies derived from the Meta-Cognitive Layer. For a given RP, let the set of stored topics be denoted $\mathcal{Z}^{RP} = \{Z_1, Z_2, \ldots, Z_u\}$,
where each topic $Z_j$ contains multiple partitions, each representing a unique or semantically distinct reasoning outcome: $Z_j = \{P_j^{(1)}, P_j^{(2)}, \ldots, P_j^{(m_j)}\}$.

When a new knowledge package $\mathcal{K}_i^{meta}$ arrives from the Meta-Cognitive Layer, its topic descriptor is first extracted and encoded into an embedding representation $\mathbf{e}_{topic} = \phi_{topic}(z_i)$, where $z_i$ is the textual or symbolic topic label associated with $\mathcal{K}_i^{meta}$ and $\phi_{topic}(\cdot)$ is the embedding model for topic encoding. Each stored topic $Z_j$ also has a representative embedding $\mathbf{e}_{Z_j}$. The semantic proximity between the new topic and an existing topic is computed using
\begin{equation}
    M_T(z_i, Z_j) = \text{STS}\bigl(\mathbf{e}_{topic}, \mathbf{e}_{Z_j}\bigr),
\end{equation}
where $STS$ denotes a semantic textual similarity operator. If the maximum similarity across all stored topics satisfies Eq.~\ref{eq:similarityIndex}
\begin{equation}\label{eq:similarityIndex}
    \max_{j} M_T(z_i, Z_j) < \theta_{topic},
\end{equation}
with $\theta_{topic}$ as the topic-matching threshold, a new topic $Z_{u+1}$ is created and appended to $\mathcal{Z}^{RP} \leftarrow \mathcal{Z}^{RP} \cup \{Z_{u+1}= \{P_{u+1}^{(1)}\}\}$. Otherwise, the new knowledge instance is associated with the closest matching topic $Z_{j^*}$, where $j^{*} = \arg\max_{j} M_T(z_i, Z_j)$.

Within the selected topic $Z_{j^*}$, the system next evaluates whether the reason or explanation associated with $H_i^{*}$ matches an existing partition. Its embedding is obtained as $\mathbf{e}_{reason} = \phi_{reason}(H_i^{*})$,  and its similarity to each stored reason embedding $\mathbf{e}_{P_{j^*}^{(k)}}$ is computed as $M_R = \text{STS}\bigl(\mathbf{e}_{reason}, \mathbf{e}_{P_{j^*}^{(k)}}\bigr)$. If the maximum similarity satisfies $\max_{k} M_R < \theta_{reason}$,
the Knowledge Layer creates a new partition $P_{j^*}^{(m_{j^*}+1)}$ within the topic $Z_{j^*}$. Otherwise, the knowledge instance reinforces the existing partition with which it best aligns, so no structural growth occurs.

As topics and partitions evolve, this layer performs adaptive reorganization to ensure coherence and reduce redundancy. When two partitions within a topic exhibit high mutual similarity, i.e.,
\begin{equation}
    \text{STS}\left(\mathbf{e}_{P_{j^*}^{(k_1)}}, \mathbf{e}_{P_{j^*}^{(k_2)}}\right) \ge \theta_{merge},
\end{equation}
they are merged into a unified representation. Conversely, if a partition exhibits significant internal semantic drift, quantified by a deviation metric $\text{DIV}(\cdot)$ exceeding a divergence threshold, the partition is split into multiple sub-partitions using Eq.~\ref{eq:subPartition}.
\begin{equation}\label{eq:subPartition}
    \text{DIV}\bigl(P_{j^*}^{(k)}\bigr) > \theta_{split}.
\end{equation}

To maintain global system consistency, each local RP periodically synchronizes its topics and partitions with global RPs through
\begin{equation}
    \mathcal{Z}^{RP}_{global} \leftarrow \text{MERGE}\bigl(\mathcal{Z}^{RP}_{global}, \mathcal{Z}^{RP}_{local}\bigr),
\end{equation}
where conflicts are resolved using similarity-based merging rules and versioning metadata.
Through these topic- and partition-level reorganizations, this layer incrementally strengthens the ReCiSt framework. This process mirrors the remodeling phase of wound healing, during which tissue is reorganized, strengthened, and integrated into surrounding structures. The resulting knowledge topology enables scalable coordination among distributed agents, supports robust context-aware decision making, and maintains long-term resilience within the DCCS.

\subsection{ReCiSt's Computational efficiency}
Although the ReCiSt framework incorporates LM- and agent-driven operations whose runtimes depend on model size, inference hardware, and system load, the algorithmic structure can still be analyzed through asymptotic complexity. Each LM invocation or prompt-driven reasoning step is treated as an oracle operation with amortized constant cost $O(1)$, since its latency does not scale with the size of the DCCS. Under this assumption, the Containment Layer (Algorithm~\ref{Layer1}) performs two dominant operations: neighbor-status probing over the adjacency set $\mathcal{N}_i$ of size $d_i$, which requires $O(d_i)$ time, and computation of the $k$-nearest neighbors $\mathcal{N}_i^{(k)}$, which requires $O(d_i \log k)$ due to distance evaluation and heap-based top-$k$ selection. Thus, the Containment Layer incurs $O(d_i \log k)$ time per monitored node. The Diagnostic Layer (Algorithm~\ref{Layer2}) parses the log file $\mathcal{L}_i$ of size $L_i$ for each failed node $N_i$ and constructs a diagnostic graph with $m_i$ extracted variables, leading to a pairwise causal evaluation cost of $O(m_i^2)$; its total complexity is therefore $O(L_i + m_i^2)$. The Meta-Cognitive Layer (Algorithm~\ref{Layer3}) generates $r_i$ reasoning micro-agents and explores $p_i$ diagnostic paths extracted from the subtrees $\Psi_i$. Since LM reasoning steps are treated as oracle operations, the complexity reduces to $O(p_i + r_i)$.
The Knowledge Layer (Algorithm~\ref{Layer4}) compares each topic embedding against $u$ stored topics in $\mathcal{Z}^{RP}$ and then evaluates the reason embedding against the $m_j$ partitions of the selected topic $Z_j$, resulting in $O(u + m_j)$ time for each knowledge update. The overall self-healing cost for a failed node $N_i$ is
$O(d_i \log k + L_i + m_i^2 + p_i + r_i + u + m_j)$, reflecting the structural complexity of the ReCiSt pipeline independent of LM inference overhead.

\section{Performance Evaluations}\label{sec:Implementation}
This section presents the experimental evaluation of the proposed \texttt{\textbf{ReCiSt}} framework. It first describes the experimental setup and performance metrics, followed by discussion of numerical results for each dataset.

\subsection{Setup}
The ReCiSt framework is implemented in Google Colab using an Intel(R) Xeon(R) CPU to represent the cloud computing environment. The LMs evaluated for ReCiSt are deployed in the cloud and accessed via the OpenAI API. The experimental setup includes geographically $k$ distributed agents among $n$ computing computing nodes. We assume that each agent $i$ ping its K-NN and subsequently initiates log acquisition from nodes that fail to respond within a specified time frame, using baseboard management controllers. Details about the computing machine, network and communications configurations are discussed along with datasets.

\subsubsection{Models}
All agents employed in the experimental evaluation were implemented using LangChain 1.0.8~\cite{langchain}. The memory module was developed as a custom repository type constructed with Pydantic~\cite{pydantic}. For embedding generation, we utilized the text-embedding-3-small model, a small embedding model developed by OpenAI~\cite{models}. To estimate semantic similarity, we adopted all-MiniLM-L6-v2, a sentence-transformer model~\cite{sts} specifically designed for semantic textual similarity tasks, enabling the computation of meaningful similarity scores between textual inputs. For our evaluation, we employ a reasoning model suited to the computational resources of the distributed computing continuum. This setting demands models capable of advanced, multi-step reasoning and capable to operate efficiently on heterogeneous and constrained computing environments. To meet these requirements, we rely on various OpenAI models~\cite{models}, including o4-mini-2025-04-16, the latest small model in the o-series, which offers strong reasoning capabilities with low latency. The gpt-5-mini-2025-08-07 model provides fast performance and high reasoning abilities. The gpt-5-nano-2025-08-07 serves as the most fast option. The gpt-5.1-2025-11-13 model for agentic tasks.

\subsubsection{Performance Metrics}
The self-healing process of ReCiSt framework mainly depend on its agents (within four layers) and their accurate and timely decision making process. So, our evaluation metrics reflect  around the effectiveness of agent operations~\cite{11270701}. In particular, the time required for self-healing, depth of analysis to avoid uncertainties, amount of micro-agents invoked during self-healing and the computational overhead for failure diagnosis, negotiation, solution discovery, and solution storage. Our evaluations further analyzed quality of agent decisions under failure conditions. This includes the rates of successful, supporting, and harmful responses, as well as the structural complexity induced by agent reasoning. These metrics are defined as follows.\\ % which focuses on analyzing the relationship between sub-tree depth and the number of micro-agent invocations to quantify the operational complexity of the self-healing process. 
% Accordingly, the performance metrics of the proposed ReCiSt is assessed using the following metrics,\\
\textbf{Time Taken for Self-Healing}: This metric measures the elapsed time from failure detection to successful recovery completing all four stages of ReCiSt pipeline.\\%, reflecting the responsiveness of the self-healing mechanism.
\textbf{CPU Consumption of the Agent}: This metric quantifies the CPU usage by an agent while performing fault diagnosis, negotiation with other agents, and self-healing process.\\
\textbf{Sub-tree Depth Complexity}: This metric show how deep the agent is started analyzing the fault.\\
\textbf{Micro-Agent Invocations}: This metric shows the  number of micro-agent calls, representing the coordination and communication overhead required for each recovery process.\\
\textbf{Quality of Decision-making}: This metrics shows the decision-making of each mic-agent in the form of acceptance, rejection, harmful or best categories defined as \textit{Accepted Rate:} The ratio of supporting responses that provide constructive and actionable guidance to the total number of responses.
\textit{Harmful Rate:}The ratio of responses that may degrade key performance indicators, such as CPU utilization, latency, or memory consumption, and negatively affect system behavior to the total number of responses.
\textit{Rejected Rate:} The ratio of responses lacking sufficient evidence or justification to the total number of responses.
\textit{Best Rate:} The ratio of responses that demonstrate strong evidentiary support, clear causal reasoning, and compliance with operational safety principles to the total number of responses.
\textit{Reasoning Depth Rate(RDR):} The ratio of system-level micro-agent invocations to the total number of instantiated micro-agents.%, derived as a function of sub-tree depth.

\subsection{Results and Analysis}\label{sec:Results}
In this section, we evaluate performance and analyze numerical results on various datasets i.e., Cloud Stateless Dataset ~\cite{8wf2-2y40-24} from IEEE DataPort and Loghub~\cite{10301257,10.1145/3650212.3652123}. %covering repository aggregates logs from distributed systems, supercomputers, server applications, and other system environments. 
The Loghub collection includes different categories of logs from systems such as ZooKeeper, Hadoop, OpenSSH and Blue Gene/L supercomputer. The subsequent sections are organized by providing dataset-wise performance evaluations and discussions. At the end, we summarize the quality of decision-making metric for all five categories of datasets.  

\subsubsection{Cloud Stateless Dataset}
\input{fig3}

The Cloud Stateless System  dataset consists of measurements collected at 5-second intervals from three cloud-based Linux virtual machines configured with 1vCPU, 1GB of memory, and a 10GB disk, all operating under a dynamically varying workload. Each record contains a timestamp alongside resource utilization and performance metrics collected through Prometheus, including cpu\_usage and memory\_usage (percentage utilization), bandwidth\_inbound and bandwidth\_outbound (throughput in GB/s or MB/s), tps (requests per second), response\_time (latency in seconds or milliseconds), and status (system status, 0 for healthy and 1 for unhealthy). Collectively, these variables offer a comprehensive temporal representation of device load, network activity, service throughput, and operational responsiveness. The failures are characterized by elevated response times and irregular bandwidth consumption, arising from network-related bottlenecks that manifest through congestion and latency spikes. These unhealthy intervals reflect performance constraints driven by fluctuations in data flow and fluctuating workload conditions, with increased CPU load appearing in some instances.

Fig.~\ref{fig:healingtimee} shows the time taken for self-healing across multiple failure instances along the dataset timeline for four models. It is noteworthy that all models successfully able to achieve self-healing, but different times. Our analysis noted that o4-mini model achieves the lowest self-recovery times, typically remaining below 300 sec ($\approx$5 min). Similarly, gpt-5.1 demonstrates low recovery times, often under 400 sec ($\approx$6–7 min), with several instances showing rapid recovery of as little as 43 sec. In contrast, gpt-5-nano' recovery times were ranged from $\approx$250 sec ($\approx$4 min) to over 1,200 sec ($\approx$20 min). The gpt-5-mini model shows higher recovery times, frequently exceeding 700–800 sec ($\approx$12–13 min), indicating slower self-healing. 
Fig.~\ref{fig:CPUusage5} depicted the CPU utilization of different models across multiple failure instances at failure time mentioned in dataset. We noted that CPU utilization remains generally stable across all models while running ReCiSt agents.  Numerically, gpt-5-mini remains around $\approx$13\%, o4-mini operates in a similar $\approx$13\% range, gpt-5.1 averages around $\approx$15\%, and gpt-5-nano close to $\approx$14\%, indicating controlled computational overhead..

Fig.~\ref{fig:microagentcalls1} and Fig.~\ref{fig:microagentcalls2} shows in the number of paths in DFS and the number of micro-agents used for self-healing. Form Fig.~\ref{fig:microagentcalls1}, we observe gpt-5.1 was the most structural complexity, which consistently generated larger sub-trees ranging from 6 to 25 paths. Nevertheless, both gpt-5.1 and o4-mini resolved failures without traversing the entire search space. This pattern highlights a strong capacity for selective path elimination and early solution detection despite an initially expansive reasoning structure. In comparison, gpt-5-nano explored sub-trees spanning $\approx$4 to 16 paths, with 4 to 25 micro-agent calls. Fig.~\ref{fig:microagentcalls2}, it is clear that the gpt-5-mini showed moderate structural complexity, generating 6 to 12 paths with 7 to 15 micro-agent calls, while o4-mini balances reasoning depth and computational overhead with 5 to 12 paths and 5 to 12 calls.

\subsubsection{Zookeper Dataset}
\input{fig4}

ZooKeeper is a centralized service that supports configuration management, naming, distributed synchronization, and group services. The log data were collected from a laboratory deployment consisting of 32 machines. Within this environment, the observed failures appeared as interrupted or broken connections.
Fig.~\ref{fig:zootime} depicted the self-healing time across two failure instances along the ZooKeeper dataset timeline for four models. It shows that gpt-5.1 achieves the lowest self-recovery times, typically ranging between 30 and 80 seconds. Also, o4-mini demonstrates slightly higher then gpt-5.1 recovery times i.e., 80-280 seconds. In contrast, gpt-5-mini shows higher recovery times, exceeding 1200 seconds, reflecting deeper but slower recovery behavior. Fig.~\ref{fig:zoocpu} illustrates the CPU utilization of different models across two failure instances at the failure times reported in the ZooKeeper dataset. We observe the CPU utilization remains stable across all models. For example, gpt-5-mini results $\approx$13\%, o4-mini operate nearly 14\% range, gpt-5.1 averages $\approx$13\%, and gpt-5-nano close to 11\%, indicating controlled computational overhead. 
From Fig.~\ref{fig:zoosub}, gpt-5.1 and gpt-5-nano exhibit similar structural complexity, with sub-trees ranging from 8 to 9 paths, whereas o4-mini shows a more compact structure. From Fig.~\ref{fig:zoocall}, we further see that gpt-5-mini incurs higher structural complexity, generating 8 to 17 paths with $\approx$10-23 micro-agent calls, while o4-mini maintains lower reasoning depth and computational overhead with 4 to 7 paths and about 1 call. gpt-5-nano also demonstrates low overhead, requiring 2-4 micro-agent calls, highlighting efficient reasoning under Zookeper dataset.

\subsubsection{Hadoop Dataset}
\input{fig5}
Hadoop is a big data processing framework. In Loghub, the Hadoop log data were from a five-node cluster operating under fault-injection scenarios, including machine outages, network disconnections, and disk full events, and their results shown in Fig.~\ref{fig:hadoop}.

From, Fig.~\ref{fig:Hadoophealingtimee} we notice the self-healing time observed across five failure instances over the Hadoop dataset timeline for the four evaluated models. The plots show that gpt-5.1 reach comparatively low recovery times, frequently remaining below 100 seconds. Also, o4-mini demonstrates fast recovery behavior, with healing times largely confined to a narrow range of approximately 60–130 seconds. In contrast, gpt-5-nano self-healing times ranging between 200 and 500 seconds, which is slightly higher than gpt-5.1 and o4-mini. Finally the gpt-5-mini result slowest among all spanning from  200 seconds to values exceeding up to 1,000 seconds. CPU utilization of Hadoop datasets and their failures are depicted in Fig.~\ref{fig:haddopCPUusage5}. The CPU usage in Hadoop dataset remains well controlled across all models. Numerically, gpt-5-nano and gpt-5-mini both exhibit average utilization of CPU is $\approx$14\%, while gpt-5.1 and o4-mini operate at marginally higher i.e., around 15\%. 
From Fig.~\ref{fig:hadooptreedepth}, we observe notable differences in failure complexity across the evaluated models. For example, gpt-5-mini shows the highest structural complexity, with sub-trees comprising $\approx$14–20 paths. In contrast, o4-mini maintains simpler structures, typically limited to about 5–10 paths. From Fig.~\ref{fig:hadoopmicroagentcalls2}, we noticed clear variation in micro-agent invocation patterns. For instance, o4-mini relies on only 1–2 micro-agent calls across the five observed failure instances. gpt-5-nano requires between 1 and 6 micro-agent calls. By comparison, gpt-5.1 invokes a larger number of micro-agent calls, reaching up to 35 in one case, suggesting an intensive reasoning process. Even though there were many ups and downs, all four models were successful in reaching a self-healing state on Hadoop dataset within ReCiSt framework.

\subsubsection{OpenSSH Dataset}
\input{fig6}

OpenSSH is a widely used tool for secure remote access based on the SSH protocol. The corresponding OpenSSH dataset contains log records related to authentication-related events and general system activities. In this dataset, possible failures are associated with broken or unsuccessful authentication attempts, including SSH authentication failures arising from brute-force attacks or mistaken login attempts. These failures and time taken to self-healing are depicted in  Fig.~\ref{fig:SSHhealingtimee}. From this, we noticed that gpt-5.1 is the fastest self-recovery agent, typically ranging between 27 and 42 seconds. Other side, o4-mini also demonstrates faster recovery gpt-5-nano and gpt-5-mini in the range of $\approx$87–107 seconds. But, gpt-5-nano and gpt-5-mini show lower recovery times, exceeding 300-800 seconds. In terms of CPU utilization as shown in Fig.~\ref{fig:SSHCPUusage5} gpt-5.1 incurs the highest CPU utilization at approximately 15\%, followed by o4-mini at around 14\%. Alternately, both gpt-5-nano and gpt-5-mini maintain lower utilization levels, close to 13\%. In terms of analysis depth and number of micro-agent calls as depicted in  Fig.~\ref{fig:SSHtreedepth} and Fig.~\ref{fig:SSHmicroagentcalls2}, respectively, gpt-5-mini shows the most complex structures, with sub-trees spanning $\approx$13–16 paths and requiring about 5–17 micro-agent calls. In contrast, o4-mini, gpt-5-nano, and gpt-5.1 demonstrate simpler recovery structures, typically involving $\approx$2–15 paths and only 1–2 micro-agent calls.

\subsubsection{Blue Gene/L (BGL) Dataset}
\input{fig7}
The Blue Gene/L (BGL) dataset is an open log collection from a supercomputing system comprising 131,072 processors and 32768 GB of memory. It contains alert messages reflecting diverse system failures, including fatal I/O errors, hardware and operating system faults, kernel-level failures, communication outages, machine-check interrupts, kernel panics, and CPU or cache-related hardware faults. 
Fig.~\ref{fig:bglhealingtimee} depicts the self-healing time across nine failure instances along the BGL dataset timeline for the evaluated models. It shows gpt-5.1 and o4-mini are the fastest self-recovery agents, taking below 250 seconds for gpt-5.1 and under 150 seconds for o4-mini. In contrast, gpt-5-nano and gpt-5-mini demonstrate slower recovery, with healing times exceeding 500 seconds and 1,200 seconds, respectively. The CPU usage of BGL data is plotted using Fig.~\ref{fig:bglCPUusage5} showing nine failure instances at the reported failure times in the BGL dataset. This confirms that gpt-5.1 and o4-mini incur slightly higher CPU utilization, $\approx$11\%, whereas gpt-5-nano and gpt-5-mini operate at lower utilization levels, close to 9\%.
From Fig.~\ref{fig:bglmicroagentcalls1} and Fig.~\ref{fig:bglmicroagentcalls}, gpt-5-mini and gpt-5.1 exhibit more complex diagnosis structures, with sub-trees spanning approximately 3–15 paths and requiring between 1 and 22 micro-agent calls. In contrast, o4-mini and gpt-5-nano demonstrate simpler diagnosis structures, typically involving approximately 2–7 DFS paths and 1–8 micro-agent calls.

\subsubsection{Quality of Agent Decisions}
We also evaluate the micro-agents' decision during the reasoning process to quantify their decision-making quality. In this process, we noted for each failure case, the rate of constructive guidance responses, the rate of responses with the potential to degrade key performance indicators and negatively affect system behavior, the rate of responses exhibiting inadequate evidentiary support, and the rate of responses demonstrating strong evidence. These evaluation further clarify causal reasoning, and adherence to operational safety principles. Further, we correlate these measurements in relation to the reasoning-depth ratio, defined as the proportion of system-level micro-agent invocations based on sub-trees depth. A summary of these results shown in Table.~\ref{tab:rate}.% illustrates the micro-agents’ reasoning outcomes across the various datasets and models. 
\input{table}

From Table.~\ref{tab:rate}, the Cloud Stateless dataset achieve supportive response in gpt-5-mini, reflected in its high acceptance rate $\approx$95\% with a full reasoning depth rate of 100\%. This superior performance in deep and consistent reasoning achieved although an higher harmful response rate of 31.8\% is noted. In contrast, gpt-5 nano shows high rejection rate $\approx$24\% and a heavy reliance on extended micro-agent-level reasoning chains $\approx$98\%, which does not translate into improved response quality $\approx$7\%. We confirm that the results indicate in this dataset are having deeper reasoning traces alone are insufficient for ensuring high-quality or safe outputs. Moreover, o4-mini demonstrates low rate of harmful outputs across several failures $\approx$5\%, alongside a high reasoning depth rate 9\% and high rate of the best solutions $\approx$53\%, shows better-calibrated internal reasoning. Similarly, for the ZooKeeper and OpenSSH datasets, there is no rejected and harmful responses resulted by all four models. Also, gpt-5-nano, gpt-5.1, and o4-mini achieving a best response rate of 100\%. However, this performance is obtained with relatively low reasoning depth (e.g., 27.9\% for gpt-5-nano and 15\% for o4-mini). This clearly show that effective recovery can be achieved without deep reasoning when fault patterns are simple. In more complex environments such as BGL and Hadoop, variability becomes more significant. For example, in BGL, gpt-5-mini balances a high accepted response rate $\approx$78.1\% with substantial reasoning depth $\approx$63.9\% and a harmful rate of 11\%. These outcomes confirm the effectiveness of feedback-driven, meta-cognitive control in balancing response quality, safety, and computational effort across diverse domains.

\section{Conclusion}\label{sec:Conclusion}
This paper proposed \texttt{\textbf{ReCiSt}}, a bio-inspired agentic self-healing framework that mapped biological wound-healing phases into autonomous containment, diagnosis, meta-cognitive reasoning, and knowledge remodeling layers to enhance resilience in DCCS. The proposed approach was implemented and tested using LM-powered agents and evaluated with multiple LMs, including gpt-5.1, gpt-5-mini, gpt-5-nano, and o4-mini, across public and heterogeneous datasets, specifically the Cloud Stateless system dataset and LogHub datasets including ZooKeeper, Hadoop, OpenSSH, and BGL. These datasets captured diverse failure types ranging from network disruptions and authentication faults to large-scale supercomputing errors. Overall, models such as gpt-5.1 and o4-mini exhibited the fastest recovery, achieving self-healing within tens of seconds, while maintaining CPU utilization below 10\%. Even though there were many ups and downs, all the models were confirming its suitability for resilient and heterogeneous computing environments. 
Primary limitation of our work is  in a controlled experimental environment using offline log datasets and cloud-based execution. Real-time factors such as live network congestion, hardware heterogeneity, mobility, security constraints, and long-running operational drift were not explicitly exercised in the current setup. In the future, our framework will be deployed and validated in real-world continuum environments and assess its robustness under live workloads and evolving failure conditions. %These deployments will enable deeper analysis of scalability, latency, and energy overheads, as well as the long-term adaptation of agent knowledge, thereby further strengthening the practical applicability of the proposed approach.

% \section*{Acknowledgments}\label{sec:acknowledges}
% This work was supported by the Research Council of Finland through the 6G Flagship program (grant 318927) and the CO2CREATION SRC project (grant 372355), by Business Finland through the Neural pub/sub research project (diary number 8754/31/2022), and by the ERDF (project numbers A81568, A91867).

\bibliographystyle{IEEEtran}
\bibliography{ref}
\balance

\end{document}

%% file: fig3.tex
\begin{figure*}[t]
	\centering
\subfloat[Self-healing time]{
	\begin{tikzpicture}
   \begin{axis}[
      width=0.45\linewidth,
      height=4cm,
      xlabel={Time of failure},xlabel style={font=\scriptsize}, xlabel style={yshift=7pt},
      ylabel={Recovery (Sec.)},ylabel style={font=\scriptsize},
      grid=both,
      xmin=1, xmax=13,
      xtick=data,
      xticklabels={
        9/8/23 1:25, 9/8/23 1:27, 9/8/23 1:27, 9/8/23 1:30, 9/8/23 1:40, 9/8/23 1:40, 9/8/23 1:42, 9/8/23 1:42, 9/8/23 1:43, 9/8/23 1:43, 9/8/23 1:43, 9/8/23 1:53, 9/8/23 1:54
    },
      xticklabel style={rotate=45, anchor=east},
      yticklabel style={rotate=45, anchor=east},
      legend columns=4,tick label style={font=\tiny},
      legend pos=north west,legend style={font=\tiny},
      % legend style={
      %   % at={(0.5,1.05)},
      %   anchor=north west,
      %   legend columns=2
      % },
      % title={Model performance across failures},
  ]
\addplot+[smooth,
    mark=*,
    error bars/.cd,
      y dir=both,
      y explicit,
]
table[
    x=idx,
    y=gpt5nano,
    y error=gpt5nano_err,
]{\levelTotal};
\addlegendentry{gpt-5-nano}

% gpt5.1
\addplot+[smooth,
    mark=square*,
    error bars/.cd,
      y dir=both,
      y explicit,
]
table[
    x=idx,
    y=gpt5,
    y error=gpt5_err,
]{\levelTotal};
\addlegendentry{gpt5.1}

% gpt5-mini
\addplot+[smooth,
    mark=triangle*,
    error bars/.cd,
      y dir=both,
      y explicit,
]
table[
    x=idx,
    y=gpt5mini,
    y error=gpt5mini_err,
]{\levelTotal};
\addlegendentry{gpt5-mini}

% o4-mini
\addplot+[smooth,
    mark=diamond*,
    error bars/.cd,
      y dir=both,
      y explicit,
]
table[
    x=idx,
    y=o4mini,
    y error=o4mini_err,
]{\levelTotal};
\addlegendentry{o4-mini}
  \end{axis}\label{fig:healingtimee}
\end{tikzpicture}
}
\subfloat[CPU Utilization]{
	\begin{tikzpicture}
   \begin{axis}[
      width=0.45\linewidth,
      height=4cm,
      xlabel={Time of Failure},xlabel style={font=\scriptsize}, xlabel style={yshift=7pt},
      ylabel={Total CPU Usage (\%)},ylabel style={font=\scriptsize},
      grid=both,
      xmin=1, xmax=13,ymax=40,
      xtick=data,
      xticklabels={
        9/8/23 1:25, 9/8/23 1:27, 9/8/23 1:27, 9/8/23 1:30, 9/8/23 1:40, 9/8/23 1:40, 9/8/23 1:42, 9/8/23 1:42, 9/8/23 1:43, 9/8/23 1:43, 9/8/23 1:43, 9/8/23 1:53, 9/8/23 1:54
    },
      xticklabel style={rotate=45, anchor=east},
      yticklabel style={rotate=45, anchor=east},
      legend columns=4,tick label style={font=\tiny},
      legend pos=north west,legend style={font=\tiny},
      % legend style={
      %   % at={(0.5,1.05)},
      %   anchor=north west,
      %   legend columns=2
      % },
      % title={Model performance across failures},
  ]
\addplot+[smooth,
    mark=*,
    error bars/.cd,
      y dir=both,
      y explicit,
]
table[
    x=idx,
    y=gpt5nano,
    y error=gpt5nano_err,
]{\cpuTotal};
\addlegendentry{gpt-5-nano}

% gpt5.1
\addplot+[smooth,
    mark=square*,
    error bars/.cd,
      y dir=both,
      y explicit,
]
table[
    x=idx,
    y=gpt5,
    y error=gpt5_err,
]{\cpuTotal};
\addlegendentry{gpt5.1}

% gpt5-mini
\addplot+[smooth,
    mark=triangle*,
    error bars/.cd,
      y dir=both,
      y explicit,
]
table[
    x=idx,
    y=gpt5mini,
    y error=gpt5mini_err,
]{\cpuTotal};
\addlegendentry{gpt5-mini}

\addplot+[smooth, mark=diamond*, error bars/.cd, y dir=both, y explicit,]
table[x=idx,  y=o4mini, y error=o4mini_err]{\cpuTotal};
\addlegendentry{o4-mini}
  \end{axis}\label{fig:CPUusage5}
\end{tikzpicture}
}\\[2pt]
\subfloat[Depth of sub-tree]{
	\begin{tikzpicture}
   \begin{axis}[
      width=0.45\linewidth,
      height=4cm,
      xlabel={Time of Failure},xlabel style={font=\scriptsize}, xlabel style={yshift=7pt},
      ylabel={Depth of sub tree},ylabel style={font=\scriptsize},
      grid=both,
      xmin=1, xmax=13,
      xtick=data,
      xticklabels={
        9/8/23 1:25, 9/8/23 1:27, 9/8/23 1:27, 9/8/23 1:30, 9/8/23 1:40, 9/8/23 1:40, 9/8/23 1:42, 9/8/23 1:42, 9/8/23 1:43, 9/8/23 1:43, 9/8/23 1:43, 9/8/23 1:53, 9/8/23 1:54
    },
      xticklabel style={rotate=45, anchor=east,font=\tiny},
      legend columns=2,tick label style={font=\scriptsize},
      legend pos=north west,legend style={font=\tiny},
  ]
    \addplot+[smooth,mark=*] table[x=idx, y=gpt5nano]{\treedepth};
    \addlegendentry{gpt-5-nano}

    \addplot+[smooth,mark=square*] table[x=idx, y=gpt5]{\treedepth};
    \addlegendentry{gpt 5.1}

    \addplot+[smooth,mark=triangle*] table[x=idx, y=gpt5mini]{\treedepth};
    \addlegendentry{gpt 5-mini}

    \addplot+[smooth,mark=diamond*] table[x=idx, y=o4mini]{\treedepth};
    \addlegendentry{o4-mini}
  \end{axis}\label{fig:microagentcalls1}
\end{tikzpicture}
}
\subfloat[Number of micro-agents called]{
	\begin{tikzpicture}
   \begin{axis}[
      width=0.45\linewidth,
      height=4cm,
      xlabel={Time of Failure},xlabel style={font=\scriptsize}, xlabel style={yshift=7pt},
      ylabel={\#micro-agent calls},ylabel style={font=\scriptsize},
      grid=both,
      xmin=1, xmax=13,
      xtick=data,
      xticklabels={
        9/8/23 1:25, 9/8/23 1:27, 9/8/23 1:27, 9/8/23 1:30, 9/8/23 1:40, 9/8/23 1:40, 9/8/23 1:42, 9/8/23 1:42, 9/8/23 1:43, 9/8/23 1:43, 9/8/23 1:43, 9/8/23 1:53, 9/8/23 1:54
    },
      xticklabel style={rotate=45, anchor=east,font=\tiny},
      legend columns=2,tick label style={font=\scriptsize},
      legend pos=north west,legend style={font=\tiny},
  ]    % One line per model
    \addplot+[smooth,mark=*] table[x=idx, y=gpt5nano]{\microagentcalls};
    \addlegendentry{gpt-5-nano}

    \addplot+[smooth,mark=square*] table[x=idx, y=gpt5]{\microagentcalls};
    \addlegendentry{gpt 5.1}

    \addplot+[smooth,mark=triangle*] table[x=idx, y=gpt5mini]{\microagentcalls};
    \addlegendentry{gpt 5-mini}

    \addplot+[smooth,mark=diamond*] table[x=idx, y=o4mini]{\microagentcalls};
    \addlegendentry{o4-mini}
  \end{axis}\label{fig:microagentcalls2}
\end{tikzpicture}
}
\caption{Performance evaluation of Cloud Stateless Dataset using the proposed ReCiSt Framework under four LMs }
    \label{fig:dataset1}%microagentcalls
\end{figure*}
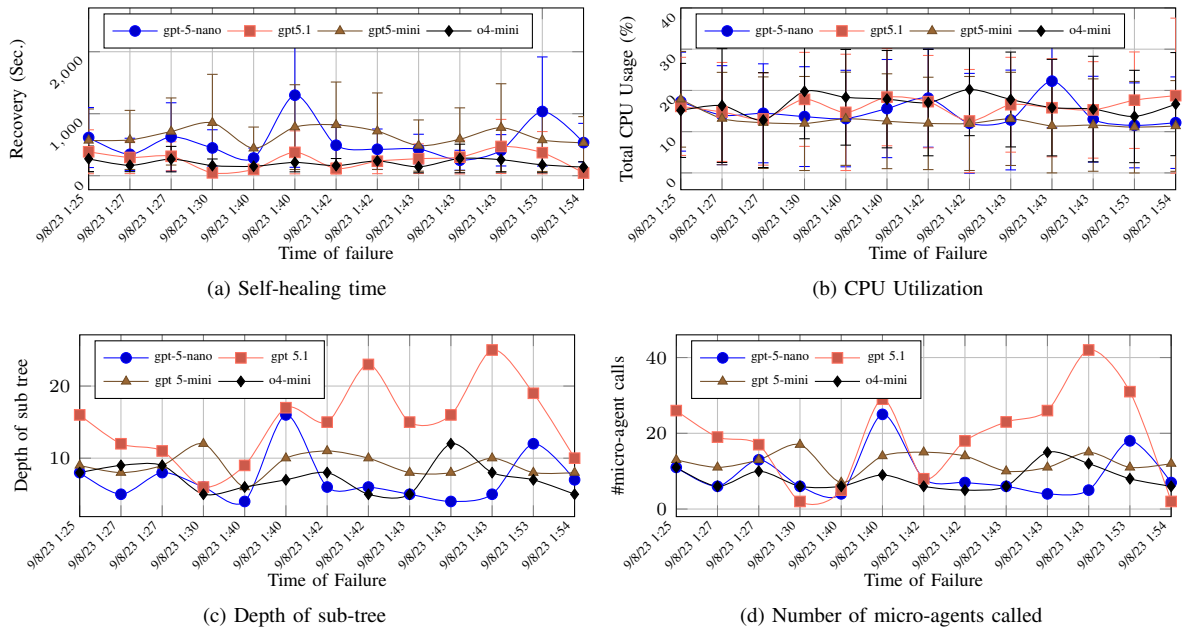

%% file: fig4.tex
\begin{figure*}[t]
	\centering
\subfloat[Self-healing time]{
\begin{tikzpicture}
\begin{axis}[
height=3.5cm,
width=5cm,
ybar,
bar width=3pt,
ymin=0,
enlarge x limits=0.35,
xtick=data,
xticklabels from table={\ZooLatency}{Failure},
xticklabel style={font=\tiny},
xlabel={Time of failure},
xlabel style={font=\scriptsize, yshift=7pt},
ylabel={Recovery (Sec.)},
ylabel style={font=\scriptsize},       yticklabel style={rotate=45, anchor=east},
ymajorgrids,
grid style={opacity=0.2},
axis line style={opacity=0.4},
tick style={opacity=0.4},
legend columns=2,tick label style={font=\tiny},
legend pos=north east,legend style={font=\tiny,
  fill opacity=0.5,  /tikz/draw=none,    
  fill=white,}, 
]
\addplot table[x=idx, y=gpt5nano] {\ZooLatency}; \addlegendentry{gpt-5-nano}
\addplot table[x=idx, y=gpt5]     {\ZooLatency}; \addlegendentry{gpt-5}
\addplot table[x=idx, y=gpt5mini] {\ZooLatency}; \addlegendentry{gpt-5-mini}
\addplot table[x=idx, y=o4mini]   {\ZooLatency}; \addlegendentry{o4-mini}
\end{axis}\label{fig:zootime}
\end{tikzpicture}
}
\subfloat[CPU Utilization]{
\begin{tikzpicture}
\begin{axis}[
height=3.5cm,
width=5cm,
ybar,
bar width=3pt,
ymin=0,
enlarge x limits=0.35,
xtick=data,
xticklabels from table={\ZooCPU}{Failure},
xticklabel style={font=\tiny},
xlabel={Time of failure},
xlabel style={font=\scriptsize, yshift=7pt},
ylabel={CPU Usage (\%)},
ylabel style={font=\scriptsize},       yticklabel style={rotate=45, anchor=east},
ymajorgrids,
grid style={opacity=0.2},
axis line style={opacity=0.4},
tick style={opacity=0.4},
legend columns=2,tick label style={font=\tiny},
legend pos=north east,legend style={font=\tiny,
  fill opacity=0.5,  /tikz/draw=none,    
  fill=white,}, ymax=30,
]
\addplot table[x=idx, y=gpt5nano] {\ZooCPU}; \addlegendentry{gpt-5-nano}
\addplot table[x=idx, y=gpt5]     {\ZooCPU}; \addlegendentry{gpt-5}
\addplot table[x=idx, y=gpt5mini] {\ZooCPU}; \addlegendentry{gpt-5-mini}
\addplot table[x=idx, y=o4mini]   {\ZooCPU}; \addlegendentry{o4-mini}
\end{axis}\label{fig:zoocpu}
\end{tikzpicture}
}
\subfloat[Depth of sub-tree]{
\begin{tikzpicture}
\begin{axis}[
height=3.5cm,
width=5cm,
ybar,
bar width=3pt,
ymin=0,
enlarge x limits=0.35,
xtick=data,
xticklabels from table={\ZooCPU}{Failure},
xticklabel style={font=\tiny},
xlabel={Time of failure},
xlabel style={font=\scriptsize, yshift=7pt},
ylabel={Subtree depth},
ylabel style={font=\scriptsize},       yticklabel style={rotate=45, anchor=east},
ymajorgrids,
grid style={opacity=0.2},
axis line style={opacity=0.4},
tick style={opacity=0.4},
legend columns=2,tick label style={font=\tiny},
legend pos=north east,legend style={font=\tiny,
  fill opacity=0.5,  /tikz/draw=none,    
  fill=white,}, ymax=30,
]
\addplot table[x=idx, y=gpt5nano] {\zootreedepth}; \addlegendentry{gpt-5-nano}
\addplot table[x=idx, y=gpt5]     {\zootreedepth}; \addlegendentry{gpt-5}
\addplot table[x=idx, y=gpt5mini] {\zootreedepth}; \addlegendentry{gpt-5-mini}
\addplot table[x=idx, y=o4mini]   {\zootreedepth}; \addlegendentry{o4-mini}
\end{axis}\label{fig:zoosub}
\end{tikzpicture}
}
\subfloat[Micro-agent calls]{
\begin{tikzpicture}
\begin{axis}[
height=3.5cm,
width=5cm,
ybar,
bar width=3pt,
ymin=0,
enlarge x limits=0.35,
xtick=data,
xticklabels from table={\zoomicroagentcalls}{Failure},
xticklabel style={font=\tiny},
xlabel={Time of failure},
xlabel style={font=\scriptsize},
ylabel={\#microagent calls},
ylabel style={font=\scriptsize},       yticklabel style={rotate=45, anchor=east},
ymajorgrids,
grid style={opacity=0.2},
axis line style={opacity=0.4},
tick style={opacity=0.4},
legend columns=2,tick label style={font=\tiny},
legend pos=north east,legend style={font=\tiny,
  fill opacity=0.5,  /tikz/draw=none,    
  fill=white,}, ymax=30,
]
\addplot table[x=idx, y=gpt5nano] {\zoomicroagentcalls}; \addlegendentry{gpt-5-nano}
\addplot table[x=idx, y=gpt5]     {\zoomicroagentcalls}; \addlegendentry{gpt-5}
\addplot table[x=idx, y=gpt5mini] {\zoomicroagentcalls}; \addlegendentry{gpt-5-mini}
\addplot table[x=idx, y=o4mini]   {\zoomicroagentcalls}; \addlegendentry{o4-mini}
\end{axis}\label{fig:zoocall}
\end{tikzpicture}
}

\caption{Performance evaluation of Zookeper dataset by the proposed Self-healing ReCiSt framework}
    \label{fig:zookeper}
\end{figure*}
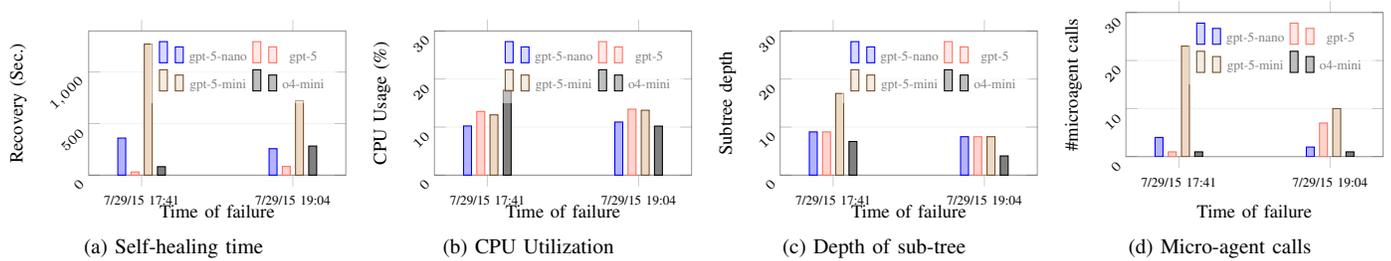

%% file: fig5.tex
\begin{figure*}[t]
	\centering
\subfloat[Self-healing time]{
	\begin{tikzpicture}
   \begin{axis}[
      width=5cm,
      height=4cm,
      xlabel={Time of failure},xlabel style={font=\scriptsize}, xlabel style={yshift=7pt},
      ylabel={Recovery (Sec.)},ylabel style={font=\scriptsize},
      grid=both,
      xmin=1, xmax=5,ymax=1700,
      xtick=data,
      xticklabels={
        2015/10/17 22:46:23, 2015/10/17 22:46:41, 2015/10/18 18:05:41, 2015/10/18 22:20:29, 2015/10/19 14:32:39
    },
      xticklabel style={rotate=90, anchor=east,font=\tiny},
      yticklabel style={font=\tiny,rotate=45, anchor=east},
      ]
\addplot+[smooth,
    mark=*,
    error bars/.cd,
      y dir=both,
      y explicit,
]
table[
    x=idx,
    y=gpt5nano,y error=gpt5nano_err,
]{\hadoopLatency};
% \addlegendentry{gpt-5-nano}

% gpt5.1
\addplot+[smooth,
    mark=square*,
    error bars/.cd,
      y dir=both,
      y explicit,
]
table[
    x=idx,
    y=gpt5, y error=gpt5_err,    
]{\hadoopLatency};
% \addlegendentry{gpt5.1}

% gpt5-mini
\addplot+[smooth,
    mark=triangle*,
    error bars/.cd,
      y dir=both,
      y explicit,
]
table[
    x=idx,
    y=gpt5mini,  y error=gpt5mini_err,  
]{\hadoopLatency};
% \addlegendentry{gpt5-mini}

% o4-mini
\addplot+[smooth,
    mark=diamond*,
    error bars/.cd,
      y dir=both,
      y explicit,
]
table[
    x=idx,
    y=o4mini,    y error=o4mini_err,  
]{\hadoopLatency};
% \addlegendentry{o4-mini}
  \end{axis}\label{fig:Hadoophealingtimee}
\end{tikzpicture}
}
\subfloat[CPU Utilization]{
	\begin{tikzpicture}
   \begin{axis}[
      width=5cm,
      height=4cm,
      xlabel={Time of failure}, xlabel style={font=\scriptsize,yshift=7pt},
      ylabel={Total CPU Usage (\%)},ylabel style={font=\scriptsize},
      grid=both,
      xmin=1, xmax=5,ymax=30,ymin=0,
      xtick=data,
      xticklabels={
        2015/10/17 22:46:23, 2015/10/17 22:46:41, 2015/10/18 18:05:41, 2015/10/18 22:20:29, 2015/10/19 14:32:39
    },
      xticklabel style={rotate=90, anchor=east},
      yticklabel style={rotate=45, anchor=east},
      legend columns=2,tick label style={font=\tiny},
      legend pos=north west,legend style={font=\tiny},
      % legend style={
      %   % at={(0.5,1.05)},
      %   anchor=north west,
      %   legend columns=2
      % },
      % title={Model performance across failures},
  ]
\addplot+[smooth,
    mark=*,
    error bars/.cd,
      y dir=both,
      y explicit,
]
table[
    x=idx,
    y=gpt5nano,y error=gpt5nano_err,
]{\hadoopCPU};
% \addlegendentry{gpt-5-nano}

% gpt5.1
\addplot+[smooth,
    mark=square*,
    error bars/.cd,
      y dir=both,
      y explicit,
]
table[
    x=idx,
    y=gpt5,y error=gpt5_err,
]{\hadoopCPU};
% \addlegendentry{gpt5.1}

% gpt5-mini
\addplot+[smooth,
    mark=triangle*,
    error bars/.cd,
      y dir=both,
      y explicit,
]
table[
    x=idx,
    y=gpt5mini,y error=gpt5mini_err,
]{\hadoopCPU};
% \addlegendentry{gpt5-mini}

% o4-mini
\addplot+[smooth,
    mark=diamond*,
    error bars/.cd,
      y dir=both,
      y explicit,
]
table[
    x=idx,
    y=o4mini,y error=o4mini_err,
]{\hadoopCPU};
% \addlegendentry{o4-mini}
  \end{axis}\label{fig:haddopCPUusage5}
\end{tikzpicture}
}
\subfloat[Depth of sub-tree]{
	\begin{tikzpicture}
  % Read the table from external file
  % \pgfplotstableread[col sep=space]{results.dat}\treedepth
   \begin{axis}[
      width=5cm,
      height=4cm,
      xlabel={Time of failure}, xlabel style={font=\scriptsize,yshift=7pt},
      ylabel={Depth of sub tree},ylabel style={font=\scriptsize},
      grid=both,
      xmin=1, xmax=5,
      xtick=data,
      xticklabels={
        2015/10/17 22:46:23, 2015/10/17 22:46:41, 2015/10/18 18:05:41, 2015/10/18 22:20:29, 2015/10/19 14:32:39
    },
      xticklabel style={rotate=90, anchor=east,font=\tiny},
      yticklabel style={font=\scriptsize},
      % legend columns=2,tick label style={font=\scriptsize},
      % legend pos=north west,legend style={font=\tiny},
  ]
    \addplot+[smooth,mark=*] table[x=idx, y=gpt5nano]{\hadooptreedepth};
    % \addlegendentry{gpt-5-nano}

    \addplot+[smooth,mark=square*] table[x=idx, y=gpt5]{\hadooptreedepth};
    % \addlegendentry{gpt 5.1}

    \addplot+[smooth,mark=triangle*] table[x=idx, y=gpt5mini]{\hadooptreedepth};
    % \addlegendentry{gpt 5-mini}

    \addplot+[smooth,mark=diamond*] table[x=idx, y=o4mini]{\hadooptreedepth};
    % \addlegendentry{o4-mini}
  \end{axis}\label{fig:hadooptreedepth}
\end{tikzpicture}
}
\subfloat[Number of micro-agents called]{
	\begin{tikzpicture}
   \begin{axis}[
      width=5cm,
      height=4cm,
      xlabel={Time of failure}, xlabel style={font=\scriptsize,yshift=7pt},
      ylabel={\#micro-agent calls},ylabel style={font=\scriptsize},
      grid=both,
      xmin=1, xmax=5,
      xtick=data,
      xticklabels={
        2015/10/17 22:46:23, 2015/10/17 22:46:41, 2015/10/18 18:05:41, 2015/10/18 22:20:29, 2015/10/19 14:32:39
    },
      xticklabel style={rotate=90, anchor=east, font=\tiny},
      legend columns=2,tick label style={font=\scriptsize},
      legend pos=north west,legend style={font=\tiny},
      % legend style={
      %   % at={(0.5,1.05)},
      %   anchor=north west,
      %   legend columns=2
      % },
      % title={Model performance across failures},
  ]    % One line per model
    \addplot+[smooth,mark=*] table[x=idx, y=gpt5nano]{\hadoopmicroagentcalls};
    \addlegendentry{gpt-5-nano}

    \addplot+[smooth,mark=square*] table[x=idx, y=gpt5]{\hadoopmicroagentcalls};
    \addlegendentry{gpt 5.1}

    \addplot+[smooth,mark=triangle*] table[x=idx, y=gpt5mini]{\hadoopmicroagentcalls};
    \addlegendentry{gpt 5-mini}

    \addplot+[smooth,mark=diamond*] table[x=idx, y=o4mini]{\hadoopmicroagentcalls};
    \addlegendentry{o4-mini}
  \end{axis}\label{fig:hadoopmicroagentcalls2}
\end{tikzpicture}
}
\caption{Performance evaluation of \textit{Hadoop} using the proposed ReCiSt Framework under four LMs }
    \label{fig:hadoop}%microagentcalls
\end{figure*}
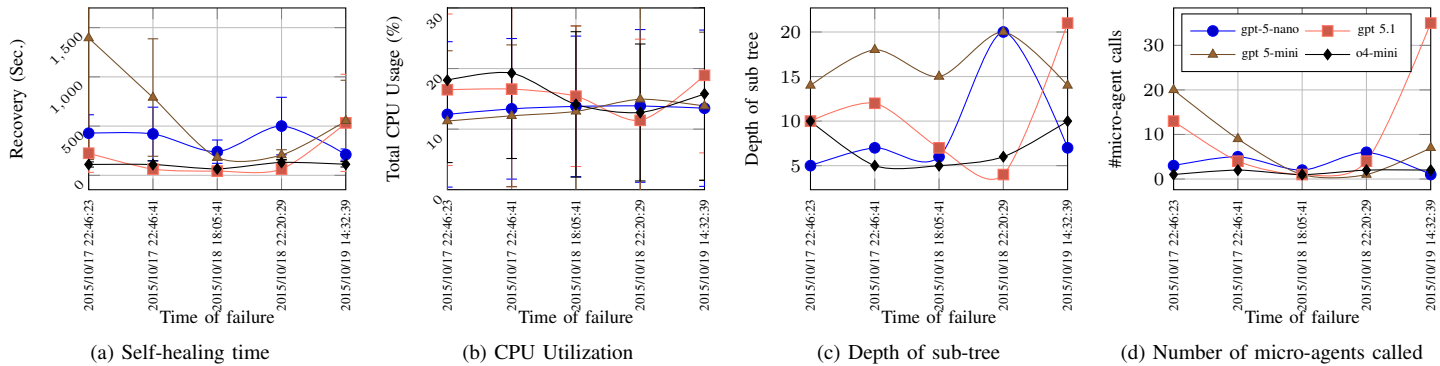

%% file: fig6.tex
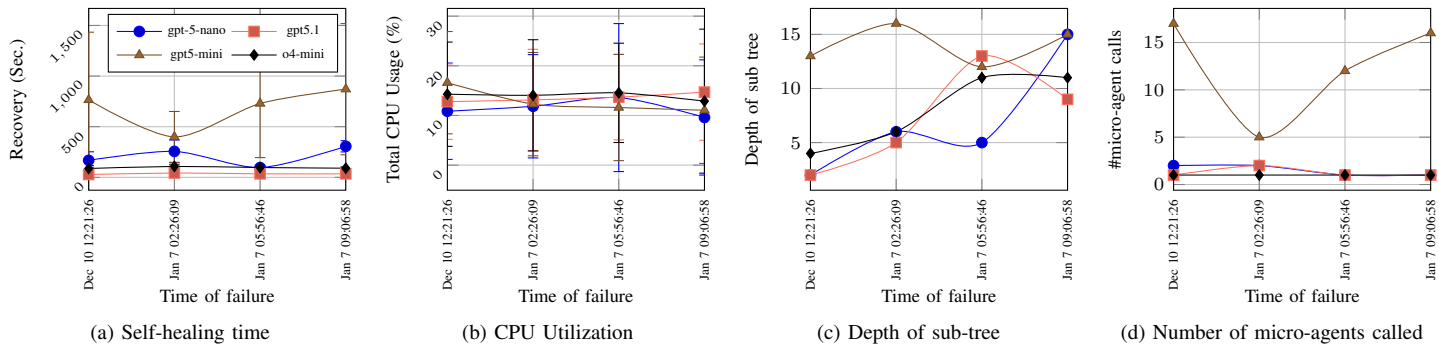
\begin{figure*}[t]
	\centering
\subfloat[Self-healing time]{
	\begin{tikzpicture}
   \begin{axis}[
      width=5cm,
      height=4cm,
      xlabel={Time of failure},xlabel style={font=\scriptsize}, xlabel style={yshift=7pt},
      ylabel={Recovery (Sec.)},ylabel style={font=\scriptsize},
      grid=both,
      xmin=1, xmax=4,%ymax=1700,
      xtick=data,
      xticklabels={
        Dec 10 12:21:26, Jan 7 02:26:09, Jan 7 05:56:46, Jan 7 09:06:58
    },
      xticklabel style={rotate=90, anchor=east,font=\tiny},
      yticklabel style={font=\tiny,rotate=45, anchor=east},
      legend columns=2,tick label style={font=\tiny},
      legend pos=north east,legend style={font=\tiny},
  ]
\addplot+[smooth,
    mark=*,
    error bars/.cd,
      y dir=both,
      y explicit,
]
table[
    x=idx,
    y=gpt5nano,y error=gpt5nano_err,
]{\sshLatency};
\addlegendentry{gpt-5-nano}

% gpt5.1
\addplot+[smooth,
    mark=square*,
    error bars/.cd,
      y dir=both,
      y explicit,
]
table[
    x=idx,
    y=gpt5, y error=gpt5_err,    
]{\sshLatency};
\addlegendentry{gpt5.1}

% gpt5-mini
\addplot+[smooth,
    mark=triangle*,
    error bars/.cd,
      y dir=both,
      y explicit,
]
table[
    x=idx,
    y=gpt5mini,  y error=gpt5mini_err,  
]{\sshLatency};
\addlegendentry{gpt5-mini}

% o4-mini
\addplot+[smooth,
    mark=diamond*,
    error bars/.cd,
      y dir=both,
      y explicit,
]
table[
    x=idx,
    y=o4mini,    y error=o4mini_err,  
]{\sshLatency};
\addlegendentry{o4-mini}
  \end{axis}\label{fig:SSHhealingtimee}
\end{tikzpicture}
}
\subfloat[CPU Utilization]{
	\begin{tikzpicture}
   \begin{axis}[
      width=5cm,
      height=4cm,
      xlabel={Time of failure}, xlabel style={font=\scriptsize,yshift=7pt},
      ylabel={Total CPU Usage (\%)},ylabel style={font=\scriptsize},
      grid=both,
      xmin=1, xmax=4,%ymax=30,ymin=0,
      xtick=data,
      xticklabels={
        Dec 10 12:21:26, Jan 7 02:26:09, Jan 7 05:56:46, Jan 7 09:06:58
    },
      xticklabel style={rotate=90, anchor=east},
      yticklabel style={rotate=45, anchor=east},
      legend columns=2,tick label style={font=\tiny},
      legend pos=north west,legend style={font=\tiny},
      % legend style={
      %   % at={(0.5,1.05)},
      %   anchor=north west,
      %   legend columns=2
      % },
      % title={Model performance across failures},
  ]
\addplot+[smooth,
    mark=*,
    error bars/.cd,
      y dir=both,
      y explicit,
]
table[
    x=idx,
    y=gpt5nano,y error=gpt5nano_err,
]{\sshCPU};
% \addlegendentry{gpt-5-nano}

% gpt5.1
\addplot+[smooth,
    mark=square*,
    error bars/.cd,
      y dir=both,
      y explicit,
]
table[
    x=idx,
    y=gpt5,y error=gpt5_err,
]{\sshCPU};
% \addlegendentry{gpt5.1}

% gpt5-mini
\addplot+[smooth,
    mark=triangle*,
    error bars/.cd,
      y dir=both,
      y explicit,
]
table[
    x=idx,
    y=gpt5mini,y error=gpt5mini_err,
]{\sshCPU};
% \addlegendentry{gpt5-mini}

% o4-mini
\addplot+[smooth,
    mark=diamond*,
    error bars/.cd,
      y dir=both,
      y explicit,
]
table[
    x=idx,
    y=o4mini,y error=o4mini_err,
]{\sshCPU};
% \addlegendentry{o4-mini}
  \end{axis}\label{fig:SSHCPUusage5}
\end{tikzpicture}
}
\subfloat[Depth of sub-tree]{
	\begin{tikzpicture}
  % Read the table from external file
  % \pgfplotstableread[col sep=space]{results.dat}\treedepth
   \begin{axis}[
      width=5cm,
      height=4cm,
      xlabel={Time of failure}, xlabel style={font=\scriptsize,yshift=7pt},
      ylabel={Depth of sub tree},ylabel style={font=\scriptsize},
      grid=both,
      xmin=1, xmax=4,
      xtick=data,
      xticklabels={
        Dec 10 12:21:26, Jan 7 02:26:09, Jan 7 05:56:46, Jan 7 09:06:58
    },
      xticklabel style={rotate=90, anchor=east,font=\tiny},
      yticklabel style={font=\scriptsize},
      % legend columns=2,tick label style={font=\scriptsize},
      % legend pos=north west,legend style={font=\tiny},
  ]
    \addplot+[smooth,mark=*] table[x=idx, y=gpt5nano]{\sshtreedepth};
    % \addlegendentry{gpt-5-nano}

    \addplot+[smooth,mark=square*] table[x=idx, y=gpt5]{\sshtreedepth};
    % \addlegendentry{gpt 5.1}

    \addplot+[smooth,mark=triangle*] table[x=idx, y=gpt5mini]{\sshtreedepth};
    % \addlegendentry{gpt 5-mini}

    \addplot+[smooth,mark=diamond*] table[x=idx, y=o4mini]{\sshtreedepth};
    % \addlegendentry{o4-mini}
  \end{axis}\label{fig:SSHtreedepth}
\end{tikzpicture}
}
\subfloat[Number of micro-agents called]{
	\begin{tikzpicture}
   \begin{axis}[
      width=5cm,
      height=4cm,
      xlabel={Time of failure}, xlabel style={font=\scriptsize,yshift=7pt},
      ylabel={\#micro-agent calls},ylabel style={font=\scriptsize},
      grid=both,
      xmin=1, xmax=4,
      xtick=data,
      xticklabels={
        Dec 10 12:21:26, Jan 7 02:26:09, Jan 7 05:56:46, Jan 7 09:06:58
    },
      xticklabel style={rotate=90, anchor=east, font=\tiny},
      tick label style={font=\scriptsize},
      % legend pos=north west,legend style={font=\tiny},
      % legend style={
      %   % at={(0.5,1.05)},
      %   anchor=north west,
      %   legend columns=2
      % },
      % title={Model performance across failures},
  ]    % One line per model
    \addplot+[smooth,mark=*] table[x=idx, y=gpt5nano]{\sshmicroagentcalls};
    % \addlegendentry{gpt-5-nano}

    \addplot+[smooth,mark=square*] table[x=idx, y=gpt5]{\sshmicroagentcalls};
    % \addlegendentry{gpt 5.1}

    \addplot+[smooth,mark=triangle*] table[x=idx, y=gpt5mini]{\sshmicroagentcalls};
    % \addlegendentry{gpt 5-mini}

    \addplot+[smooth,mark=diamond*] table[x=idx, y=o4mini]{\sshmicroagentcalls};
    % \addlegendentry{o4-mini}
  \end{axis}\label{fig:SSHmicroagentcalls2}
\end{tikzpicture}
}
\caption{Performance evaluation of \textit{OpenSSH} using the proposed ReCiSt Framework under four LMs }
    \label{fig:openSSH}%microagentcalls
\end{figure*}

%% file: fig7.tex
\begin{figure*}[t]
	\centering
\subfloat[Self-healing time]{
	\begin{tikzpicture}
   \begin{axis}[
      width=0.45\linewidth,
      height=4cm,
      xlabel={Time of failure},xlabel style={font=\scriptsize}, xlabel style={yshift=7pt},
      ylabel={Recovery (Sec.)},ylabel style={font=\scriptsize},
      grid=both,
      xmin=1, xmax=9,
      xtick=data,
      xticklabels={
        2005-06-04-04.45.30, 2005-06-14-09.49.09, 2005-12-16-14.59.02, 2005-12-24-12.41.59, 2005-12-24-16.45.51, 2005-12-24-16.46.10, 2005-12-25-22.06.55, 2005-12-25-22.07.07, 2006-01-04-08.00.05 
    },
      xticklabel style={rotate=45, anchor=east},
      yticklabel style={rotate=45, anchor=east},
      legend columns=4,tick label style={font=\tiny},
      legend pos=north west,legend style={font=\tiny},
      % legend style={
      %   % at={(0.5,1.05)},
      %   anchor=north west,
      %   legend columns=2
      % },
      % title={Model performance across failures},
  ]
\addplot+[smooth,
    mark=*,
    error bars/.cd,
      y dir=both,
      y explicit,
]
table[
    x=idx,
    y=gpt5nano,
    y error=gpt5nano_err,
]{\bglLatency};
\addlegendentry{gpt-5-nano}

% gpt5.1
\addplot+[smooth,
    mark=square*,
    error bars/.cd,
      y dir=both,
      y explicit,
]
table[
    x=idx,
    y=gpt5,
    y error=gpt5_err,
]{\bglLatency};
\addlegendentry{gpt5.1}

% gpt5-mini
\addplot+[smooth,
    mark=triangle*,
    error bars/.cd,
      y dir=both,
      y explicit,
]
table[
    x=idx,
    y=gpt5mini,
    y error=gpt5mini_err,
]{\bglLatency};
\addlegendentry{gpt5-mini}

% o4-mini
\addplot+[smooth,
    mark=diamond*,
    error bars/.cd,
      y dir=both,
      y explicit,
]
table[
    x=idx,
    y=o4mini,
    y error=o4mini_err,
]{\bglLatency};
\addlegendentry{o4-mini}
  \end{axis}\label{fig:bglhealingtimee}
\end{tikzpicture}
}
\subfloat[CPU Utilization]{
	\begin{tikzpicture}
   \begin{axis}[
      width=0.45\linewidth,
      height=4cm,
      xlabel={Time of Failure},xlabel style={font=\scriptsize}, xlabel style={yshift=7pt},
      ylabel={Total CPU Usage (\%)},ylabel style={font=\scriptsize},
      grid=both,
      xmin=1, xmax=9, %ymax=120,
      xtick=data,
      xticklabels={
        2005-06-04-04.45.30, 2005-06-14-09.49.09, 2005-12-16-14.59.02, 2005-12-24-12.41.59, 2005-12-24-16.45.51, 2005-12-24-16.46.10, 2005-12-25-22.06.55, 2005-12-25-22.07.07, 2006-01-04-08.00.05
    },
      xticklabel style={rotate=45, anchor=east},
      yticklabel style={rotate=45, anchor=east},
      legend columns=4,tick label style={font=\tiny},
      legend pos=north west,legend style={font=\tiny},
      % legend style={
      %   % at={(0.5,1.05)},
      %   anchor=north west,
      %   legend columns=2
      % },
      % title={Model performance across failures},
  ]
\addplot+[smooth,
    mark=*,
    error bars/.cd,
      y dir=both,
      y explicit,
]
table[
    x=idx,
    y=gpt5nano,
    y error=gpt5nano_err,
]{\bglCPU};
\addlegendentry{gpt-5-nano}

% gpt5.1
\addplot+[smooth,
    mark=square*,
    error bars/.cd,
      y dir=both,
      y explicit,
]
table[
    x=idx,
    y=gpt5,
    y error=gpt5_err,
]{\bglCPU};
\addlegendentry{gpt5.1}

% gpt5-mini
\addplot+[smooth,
    mark=triangle*,
    error bars/.cd,
      y dir=both,
      y explicit,
]
table[
    x=idx,
    y=gpt5mini,
    y error=gpt5mini_err,
]{\bglCPU};
\addlegendentry{gpt5-mini}

% o4-mini
\addplot+[smooth,
    mark=diamond*,
    error bars/.cd,
      y dir=both,
      y explicit,
]
table[
    x=idx,
    y=o4mini,
    y error=o4mini_err,
]{\bglCPU};
\addlegendentry{o4-mini}
  \end{axis}\label{fig:bglCPUusage5}
\end{tikzpicture}
}\\[2pt]
\subfloat[Depth of sub-tree]{
	\begin{tikzpicture}
  % Read the table from external file
  % \pgfplotstableread[col sep=space]{results.dat}\treedepth
   \begin{axis}[
      width=0.45\linewidth,
      height=4cm,
      xlabel={Time of failure},xlabel style={font=\scriptsize}, xlabel style={yshift=7pt},
      ylabel={Depth of sub tree},ylabel style={font=\scriptsize},
      grid=both,
      xmin=1, xmax=9,
      xtick=data,
      xticklabels={
        2005-06-04-04.45.30, 2005-06-14-09.49.09, 2005-12-16-14.59.02, 2005-12-24-12.41.59, 2005-12-24-16.45.51, 2005-12-24-16.46.10, 2005-12-25-22.06.55, 2005-12-25-22.07.07, 2006-01-04-08.00.05
    },
      xticklabel style={rotate=45, anchor=east,font=\tiny},
      legend columns=2,tick label style={font=\scriptsize},
      legend pos=north west,legend style={font=\tiny},
  ]
    \addplot+[smooth,mark=*] table[x=idx, y=gpt5nano]{\bgltreedepth};
    \addlegendentry{gpt-5-nano}

    \addplot+[smooth,mark=square*] table[x=idx, y=gpt5]{\bgltreedepth};
    \addlegendentry{gpt 5.1}

    \addplot+[smooth,mark=triangle*] table[x=idx, y=gpt5mini]{\bgltreedepth};
    \addlegendentry{gpt 5-mini}

    \addplot+[smooth,mark=diamond*] table[x=idx, y=o4mini]{\bgltreedepth};
    \addlegendentry{o4-mini}
  \end{axis}\label{fig:bglmicroagentcalls1}
\end{tikzpicture}
}
\subfloat[Number of micro-agents called]{
	\begin{tikzpicture}
   \begin{axis}[
      width=0.45\linewidth,
      height=4cm,
      xlabel={Time of failure},xlabel style={font=\scriptsize}, xlabel style={yshift=7pt},
      ylabel={\#micro-agent calls},ylabel style={font=\scriptsize},
      grid=both,
      xmin=1, xmax=9,
      xtick=data,
      xticklabels={
        2005-06-04-04.45.30, 2005-06-14-09.49.09, 2005-12-16-14.59.02, 2005-12-24-12.41.59, 2005-12-24-16.45.51, 2005-12-24-16.46.10, 2005-12-25-22.06.55, 2005-12-25-22.07.07, 2006-01-04-08.00.05
    },
      xticklabel style={rotate=45, anchor=east,font=\tiny},
      legend columns=2,tick label style={font=\scriptsize},
      legend pos=north west,legend style={font=\tiny},
      % legend style={
      %   % at={(0.5,1.05)},
      %   anchor=north west,
      %   legend columns=2
      % },
      % title={Model performance across failures},
  ]    % One line per model
    \addplot+[smooth,mark=*] table[x=idx, y=gpt5nano]{\bglmicroagentcalls};
    \addlegendentry{gpt-5-nano}

    \addplot+[smooth,mark=square*] table[x=idx, y=gpt5]{\bglmicroagentcalls};
    \addlegendentry{gpt 5.1}

    \addplot+[smooth,mark=triangle*] table[x=idx, y=gpt5mini]{\bglmicroagentcalls};
    \addlegendentry{gpt 5-mini}

    \addplot+[smooth,mark=diamond*] table[x=idx, y=o4mini]{\bglmicroagentcalls};
    \addlegendentry{o4-mini}
  \end{axis}\label{fig:bglmicroagentcalls}
\end{tikzpicture}
}
\caption{Performance evaluation of BGL dataset using the proposed ReCiSt Framework under four LMs }
    \label{fig:BGL}%microagentcalls
\end{figure*}
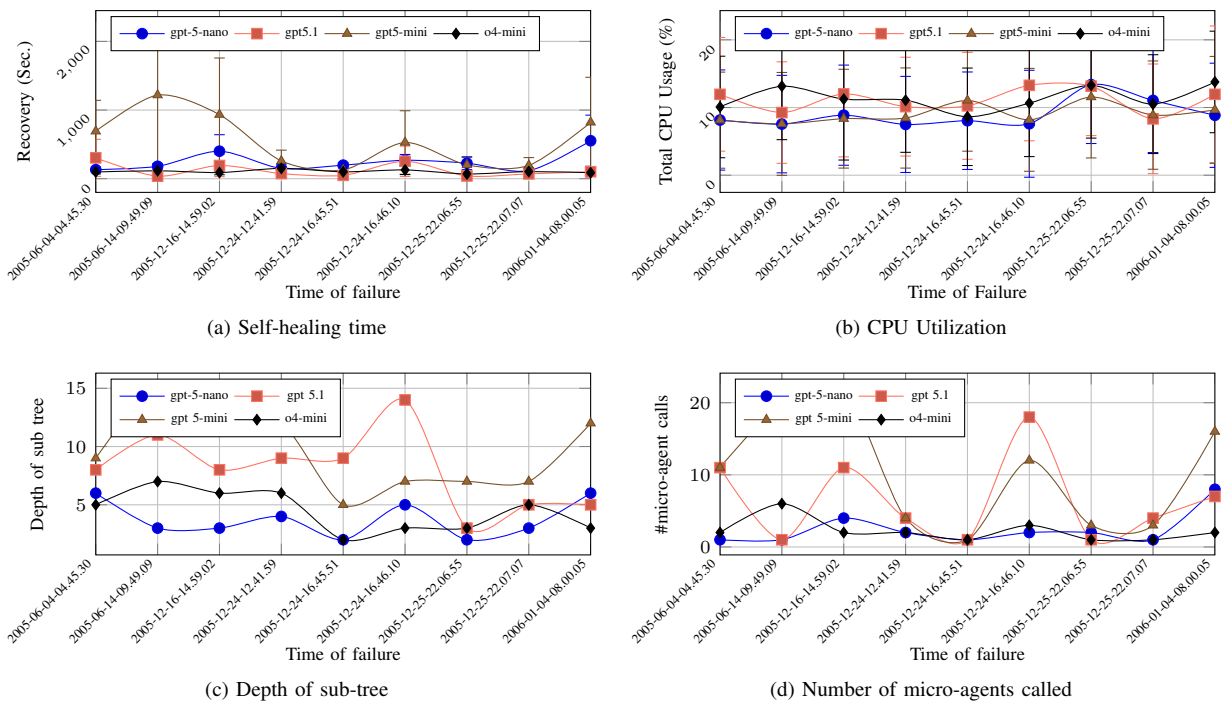

%% file: table.tex
\begin{table}[t]
\centering
\caption{Quality of decision making of agents' responses across five datasets and four models}
\label{tab:rate}
\resizebox{!}{0.12\paperheight}{%
\begin{tabular}{
>{\columncolor[HTML]{FFFFFF}}c 
>{\columncolor[HTML]{FFFFFF}}l 
>{\columncolor[HTML]{FFFFFF}}c 
>{\columncolor[HTML]{FFFFFF}}c 
>{\columncolor[HTML]{FFFFFF}}c 
>{\columncolor[HTML]{FFFFFF}}c 
>{\columncolor[HTML]{FFFFFF}}c }
\hline
\multicolumn{1}{c|}{\cellcolor[HTML]{FFFFFF}\textbf{Dataset}} &
  \multicolumn{1}{c|}{\cellcolor[HTML]{FFFFFF}\textbf{Model name}} &
  \multicolumn{1}{c|}{\cellcolor[HTML]{FFFFFF}\textbf{Best}} &
  \multicolumn{1}{c|}{\cellcolor[HTML]{FFFFFF}\textbf{Accepted}} &
  \multicolumn{1}{c|}{\cellcolor[HTML]{FFFFFF}\textbf{Rejected}} &
  \multicolumn{1}{c|}{\cellcolor[HTML]{FFFFFF}\textbf{Harmful}} &
  \textbf{RDR} \\ \hline
\multicolumn{1}{c|}{\cellcolor[HTML]{FFFFFF}} &
  \multicolumn{1}{l|}{\cellcolor[HTML]{FFFFFF}\textbf{gpt 5 nano}} &
  \multicolumn{1}{c|}{\cellcolor[HTML]{FFFFFF}0.077} &
  \multicolumn{1}{c|}{\cellcolor[HTML]{FFFFFF}0.753} &
  \multicolumn{1}{c|}{\cellcolor[HTML]{FFFFFF}0.244} &
  \multicolumn{1}{c|}{\cellcolor[HTML]{FFFFFF}0.077} &
  0.986 \\ \cline{2-7} 
\multicolumn{1}{c|}{\cellcolor[HTML]{FFFFFF}} &
  \multicolumn{1}{l|}{\cellcolor[HTML]{FFFFFF}\textbf{gpt 5.1}} &
  \multicolumn{1}{c|}{\cellcolor[HTML]{FFFFFF}0.462} &
  \multicolumn{1}{c|}{\cellcolor[HTML]{FFFFFF}0.733} &
  \multicolumn{1}{c|}{\cellcolor[HTML]{FFFFFF}0.162} &
  \multicolumn{1}{c|}{\cellcolor[HTML]{FFFFFF}0.154} &
  0.728 \\ \cline{2-7} 
\multicolumn{1}{c|}{\cellcolor[HTML]{FFFFFF}} &
  \multicolumn{1}{l|}{\cellcolor[HTML]{FFFFFF}\textbf{gpt 5 mini}} &
  \multicolumn{1}{c|}{\cellcolor[HTML]{FFFFFF}0} &
  \multicolumn{1}{c|}{\cellcolor[HTML]{FFFFFF}0.955} &
  \multicolumn{1}{c|}{\cellcolor[HTML]{FFFFFF}0.051} &
  \multicolumn{1}{c|}{\cellcolor[HTML]{FFFFFF}0.318} &
  1 \\ \cline{2-7} 
\multicolumn{1}{c|}{\multirow{-4}{*}{\cellcolor[HTML]{FFFFFF}\textbf{\begin{tabular}[c]{@{}c@{}}Cloud\\ Stateless\end{tabular}}}} &
  \multicolumn{1}{l|}{\cellcolor[HTML]{FFFFFF}\textbf{o4 mini}} &
  \multicolumn{1}{c|}{\cellcolor[HTML]{FFFFFF}0.538} &
  \multicolumn{1}{c|}{\cellcolor[HTML]{FFFFFF}0.628} &
  \multicolumn{1}{c|}{\cellcolor[HTML]{FFFFFF}0.301} &
  \multicolumn{1}{c|}{\cellcolor[HTML]{FFFFFF}0.055} &
  0.905 \\ \hline
 &
   &
   &
   &
   &
   &
   \\ \hline
\multicolumn{1}{c|}{\cellcolor[HTML]{FFFFFF}{\color[HTML]{000000} }} &
  \multicolumn{1}{l|}{\cellcolor[HTML]{FFFFFF}{\color[HTML]{1F2328} \textbf{gpt 5 nano}}} &
  \multicolumn{1}{c|}{\cellcolor[HTML]{FFFFFF}1} &
  \multicolumn{1}{c|}{\cellcolor[HTML]{FFFFFF}0.625} &
  \multicolumn{1}{c|}{\cellcolor[HTML]{FFFFFF}0} &
  \multicolumn{1}{c|}{\cellcolor[HTML]{FFFFFF}0} &
  0.279 \\ \cline{2-7} 
\multicolumn{1}{c|}{\cellcolor[HTML]{FFFFFF}{\color[HTML]{000000} }} &
  \multicolumn{1}{l|}{\cellcolor[HTML]{FFFFFF}\textbf{gpt 5.1}} &
  \multicolumn{1}{c|}{\cellcolor[HTML]{FFFFFF}1} &
  \multicolumn{1}{c|}{\cellcolor[HTML]{FFFFFF}0.429} &
  \multicolumn{1}{c|}{\cellcolor[HTML]{FFFFFF}0} &
  \multicolumn{1}{c|}{\cellcolor[HTML]{FFFFFF}0} &
  0.392 \\ \cline{2-7} 
\multicolumn{1}{c|}{\cellcolor[HTML]{FFFFFF}{\color[HTML]{000000} }} &
  \multicolumn{1}{l|}{\cellcolor[HTML]{FFFFFF}\textbf{gpt 5 mini}} &
  \multicolumn{1}{c|}{\cellcolor[HTML]{FFFFFF}0.5} &
  \multicolumn{1}{c|}{\cellcolor[HTML]{FFFFFF}0.978} &
  \multicolumn{1}{c|}{\cellcolor[HTML]{FFFFFF}0} &
  \multicolumn{1}{c|}{\cellcolor[HTML]{FFFFFF}0} &
  0.942 \\ \cline{2-7} 
\multicolumn{1}{c|}{\multirow{-4}{*}{\cellcolor[HTML]{FFFFFF}{\color[HTML]{000000} \textbf{Zookeeper}}}} &
  \multicolumn{1}{l|}{\cellcolor[HTML]{FFFFFF}\textbf{o4 mini}} &
  \multicolumn{1}{c|}{\cellcolor[HTML]{FFFFFF}1} &
  \multicolumn{1}{c|}{\cellcolor[HTML]{FFFFFF}0} &
  \multicolumn{1}{c|}{\cellcolor[HTML]{FFFFFF}0} &
  \multicolumn{1}{c|}{\cellcolor[HTML]{FFFFFF}0} &
  0.15 \\ \hline
\textbf{} &
  \textbf{} &
   &
   &
   &
   &
   \\ \hline
\multicolumn{1}{c|}{\cellcolor[HTML]{FFFFFF}} &
  \multicolumn{1}{l|}{\cellcolor[HTML]{FFFFFF}\textbf{gpt 5 nano}} &
  \multicolumn{1}{c|}{\cellcolor[HTML]{FFFFFF}1} &
  \multicolumn{1}{c|}{\cellcolor[HTML]{FFFFFF}0.25} &
  \multicolumn{1}{c|}{\cellcolor[HTML]{FFFFFF}0} &
  \multicolumn{1}{c|}{\cellcolor[HTML]{FFFFFF}0} &
  0.288 \\ \cline{2-7} 
\multicolumn{1}{c|}{\cellcolor[HTML]{FFFFFF}} &
  \multicolumn{1}{l|}{\cellcolor[HTML]{FFFFFF}\textbf{gpt 5.1}} &
  \multicolumn{1}{c|}{\cellcolor[HTML]{FFFFFF}1} &
  \multicolumn{1}{c|}{\cellcolor[HTML]{FFFFFF}0.125} &
  \multicolumn{1}{c|}{\cellcolor[HTML]{FFFFFF}0} &
  \multicolumn{1}{c|}{\cellcolor[HTML]{FFFFFF}0} &
  0.195 \\ \cline{2-7} 
\multicolumn{1}{c|}{\cellcolor[HTML]{FFFFFF}} &
  \multicolumn{1}{l|}{\cellcolor[HTML]{FFFFFF}\textbf{gpt 5 mini}} &
  \multicolumn{1}{c|}{\cellcolor[HTML]{FFFFFF}0.75} &
  \multicolumn{1}{c|}{\cellcolor[HTML]{FFFFFF}0.914} &
  \multicolumn{1}{c|}{\cellcolor[HTML]{FFFFFF}0} &
  \multicolumn{1}{c|}{\cellcolor[HTML]{FFFFFF}0} &
  0.733 \\ \cline{2-7} 
\multicolumn{1}{c|}{\multirow{-4}{*}{\cellcolor[HTML]{FFFFFF}\textbf{OpenSSH}}} &
  \multicolumn{1}{l|}{\cellcolor[HTML]{FFFFFF}\textbf{o4 mini}} &
  \multicolumn{1}{c|}{\cellcolor[HTML]{FFFFFF}1} &
  \multicolumn{1}{c|}{\cellcolor[HTML]{FFFFFF}0.125} &
  \multicolumn{1}{c|}{\cellcolor[HTML]{FFFFFF}0} &
  \multicolumn{1}{c|}{\cellcolor[HTML]{FFFFFF}0} &
  0.170 \\ \hline
\textbf{} &
  \textbf{} &
   &
   &
   &
   &
   \\ \hline
\multicolumn{1}{c|}{\cellcolor[HTML]{FFFFFF}} &
  \multicolumn{1}{l|}{\cellcolor[HTML]{FFFFFF}\textbf{gpt 5 nano}} &
  \multicolumn{1}{c|}{\cellcolor[HTML]{FFFFFF}0.778} &
  \multicolumn{1}{c|}{\cellcolor[HTML]{FFFFFF}0.431} &
  \multicolumn{1}{c|}{\cellcolor[HTML]{FFFFFF}0} &
  \multicolumn{1}{c|}{\cellcolor[HTML]{FFFFFF}0} &
  0.499 \\ \cline{2-7} 
\multicolumn{1}{c|}{\cellcolor[HTML]{FFFFFF}} &
  \multicolumn{1}{l|}{\cellcolor[HTML]{FFFFFF}\textbf{gpt 5.1}} &
  \multicolumn{1}{c|}{\cellcolor[HTML]{FFFFFF}0.556} &
  \multicolumn{1}{c|}{\cellcolor[HTML]{FFFFFF}0.611} &
  \multicolumn{1}{c|}{\cellcolor[HTML]{FFFFFF}0} &
  \multicolumn{1}{c|}{\cellcolor[HTML]{FFFFFF}0} &
  0.569 \\ \cline{2-7} 
\multicolumn{1}{c|}{\cellcolor[HTML]{FFFFFF}} &
  \multicolumn{1}{l|}{\cellcolor[HTML]{FFFFFF}\textbf{gpt 5 mini}} &
  \multicolumn{1}{c|}{\cellcolor[HTML]{FFFFFF}0.556} &
  \multicolumn{1}{c|}{\cellcolor[HTML]{FFFFFF}0.781} &
  \multicolumn{1}{c|}{\cellcolor[HTML]{FFFFFF}0} &
  \multicolumn{1}{c|}{\cellcolor[HTML]{FFFFFF}0.111} &
  0.639 \\ \cline{2-7} 
\multicolumn{1}{c|}{\multirow{-4}{*}{\cellcolor[HTML]{FFFFFF}\textbf{BGL}}} &
  \multicolumn{1}{l|}{\cellcolor[HTML]{FFFFFF}\textbf{o4 mini}} &
  \multicolumn{1}{c|}{\cellcolor[HTML]{FFFFFF}1} &
  \multicolumn{1}{c|}{\cellcolor[HTML]{FFFFFF}0.315} &
  \multicolumn{1}{c|}{\cellcolor[HTML]{FFFFFF}0.222} &
  \multicolumn{1}{c|}{\cellcolor[HTML]{FFFFFF}0.111} &
  0.402 \\ \hline
\textbf{} &
  \textbf{} &
   &
   &
   &
   &
   \\ \hline
\multicolumn{1}{c|}{\cellcolor[HTML]{FFFFFF}} &
  \multicolumn{1}{l|}{\cellcolor[HTML]{FFFFFF}\textbf{gpt 5 nano}} &
  \multicolumn{1}{c|}{\cellcolor[HTML]{FFFFFF}1} &
  \multicolumn{1}{c|}{\cellcolor[HTML]{FFFFFF}0.56} &
  \multicolumn{1}{c|}{\cellcolor[HTML]{FFFFFF}0} &
  \multicolumn{1}{c|}{\cellcolor[HTML]{FFFFFF}0} &
  0.315 \\ \cline{2-7} 
\multicolumn{1}{c|}{\cellcolor[HTML]{FFFFFF}} &
  \multicolumn{1}{l|}{\cellcolor[HTML]{FFFFFF}\textbf{gpt 5.1}} &
  \multicolumn{1}{c|}{\cellcolor[HTML]{FFFFFF}0.6} &
  \multicolumn{1}{c|}{\cellcolor[HTML]{FFFFFF}0.7} &
  \multicolumn{1}{c|}{\cellcolor[HTML]{FFFFFF}0} &
  \multicolumn{1}{c|}{\cellcolor[HTML]{FFFFFF}0} &
  0.622 \\ \cline{2-7} 
\multicolumn{1}{c|}{\cellcolor[HTML]{FFFFFF}} &
  \multicolumn{1}{l|}{\cellcolor[HTML]{FFFFFF}\textbf{gpt 5 mini}} &
  \multicolumn{1}{c|}{\cellcolor[HTML]{FFFFFF}0.8} &
  \multicolumn{1}{c|}{\cellcolor[HTML]{FFFFFF}0.539} &
  \multicolumn{1}{c|}{\cellcolor[HTML]{FFFFFF}0.2} &
  \multicolumn{1}{c|}{\cellcolor[HTML]{FFFFFF}0} &
  0.349 \\ \cline{2-7} 
\multicolumn{1}{c|}{\multirow{-4}{*}{\cellcolor[HTML]{FFFFFF}\textbf{Hadoop}}} &
  \multicolumn{1}{l|}{\cellcolor[HTML]{FFFFFF}\textbf{o4 mini}} &
  \multicolumn{1}{c|}{\cellcolor[HTML]{FFFFFF}1} &
  \multicolumn{1}{c|}{\cellcolor[HTML]{FFFFFF}0.3} &
  \multicolumn{1}{c|}{\cellcolor[HTML]{FFFFFF}0} &
  \multicolumn{1}{c|}{\cellcolor[HTML]{FFFFFF}0} &
  0.187 \\ \hline
\end{tabular}
}
\end{table}

%% file: ref.bib
@ARTICLE{donta2025human,
  author={Donta, Praveen Kumar and Sedlak, Boris and Murturi, Ilir and Pujol, Victor Casamayor and Dustdar, Schahram},
  journal={IEEE Internet Computing}, 
  title={Human-Based Distributed Intelligence in Computing Continuum Systems}, 
  year={2025},
  volume={29},
  number={2},
  pages={61-68},
  doi={10.1109/MIC.2024.3460908}
}

@misc{alves20256gresiliencewhite,
      title={{6G} Resilience -- White Paper}, 
      author={Hirley Alves and others},
      year={2025},
      eprint={2509.09005},
      archivePrefix={arXiv},
      primaryClass={eess.SP},
      url={https://arxiv.org/abs/2509.09005}, 
}

@ARTICLE{dustdar2022distributed,
  author={Dustdar, Schahram and Pujol, Victor Casamayor and Donta, Praveen Kumar},
  journal={IEEE Transactions on Knowledge and Data Engineering}, 
  title={On Distributed Computing Continuum Systems}, 
  year={2023},
  volume={35},
  number={4},
  pages={4092-4105},
  doi={10.1109/TKDE.2022.3142856}}

@ARTICLE{9466159,
  author={Nikolić, Jovan and Jubatyrov, Nursultan and Pournaras, Evangelos},
  journal={IEEE Trans. on Netw. Serv. Manag..}, 
  title={Self-Healing Dilemmas in Distributed Systems: Fault Correction vs. Fault Tolerance}, 
  year={2021},
  volume={18},
  number={3},
  pages={2728-2741},
  doi={10.1109/TNSM.2021.3092939}}

@article{donta2023governance,
  title={Governance and sustainability of distributed continuum systems: A big data approach},
  author={Donta, Praveen Kumar and Sedlak, Boris and Casamayor Pujol, Victor and Dustdar, Schahram},
  journal={Journal of Big Data},
  volume={10},
  number={1},
  pages={53},
  year={2023},
  publisher={Springer},
  url= {https://doi.org/10.1186/s40537-023-00737-0},
}

@article{10.1145/3774946,
author = {Saleh, Alaa and Tarkoma, Sasu and Lindgren, Anders and Donta, Praveen Kumar and Dustdar, Schahram and Pirttikangas, Susanna and Lov\'{e}n, Lauri},
title = {{MemIndex}: Agentic Event-based Distributed Memory Management for Multi-agent Systems},
year = {2025},
publisher = {ACM},
address = {New York, NY, USA},
issn = {1556-4665},
url = {https://doi.org/10.1145/3774946},
doi = {10.1145/3774946},
note = {Just Accepted},
journal = {ACM Trans. Auton. Adapt. Syst.},
month = nov
}

@article{enoch2008basic,
  title={Basic science of wound healing},
  author={Enoch, Stuart and Leaper, David John},
  journal={Surgery (oxford)},
  volume={26},
  number={2},
  pages={31--37},
  year={2008},
  url={https://doi.org/10.1383/surg.23.2.37.60352},
  publisher={Elsevier}
}

@ARTICLE{10562327,
  author={Fang, Honglin and Yu, Peng and Tan, Can and Zhang, Junye and Lin, Dahua and Zhang, Liyan and Zhang, Yong and Li, Wenjing and Meng, Luoming},
  journal={IEEE Network}, 
  title={Self-Healing in Knowledge-Driven Autonomous Networks: Context, Challenges, and Future Directions}, 
  year={2024},
  volume={38},
  number={6},
  pages={425-432},
  doi={10.1109/MNET.2024.3416850}}

@INPROCEEDINGS{diaz2023akats,
  author={Diaz-de-Arcaya, Josu and Torre-Bastida, Ana I. and Bonilla, Lander and López-de-Armentia, Juan and Miñón, Raúl and Zarate, Gorka and Almeida, Aitor},
  booktitle={2023 8th International Conference on Smart and Sustainable Technologies (SpliTech)}, 
  title={Akats: A System for Resilient Deployments on Edge Computing Environments Using Federated Machine Learning Techniques}, 
  year={2023},
  volume={},
  number={},
  pages={1-4},
  doi={10.23919/SpliTech58164.2023.10193302}}

@INPROCEEDINGS{sen2021resilient,
  author={Sen, Tanmoy and Shen, Haiying and Saad, Walid and Doan, Thinh},
  booktitle={2021 IEEE 18th International Conference on Mobile Ad Hoc and Smart Systems}, 
  title={A Resilient and Robust Edge-Cloud Network System Supporting {CPS}}, 
  year={2021},
  volume={},
  number={},
  pages={234-242},
  doi={10.1109/MASS52906.2021.00039}}

@article{altaweel2022rsock,
author = {Altaweel, Ala and Yang, Chen and Stoleru, Radu and Bhunia, Suman and Sagor, Mohammad and Maurice, Maxwell and Blalock, Roger},
title = {{RSock}: A resilient routing protocol for mobile Fog/Edge networks},
year = {2022},
issue_date = {Sep 2022},
publisher = {Elsevier Science Publishers B. V.},
address = {NLD},
volume = {134},
number = {C},
issn = {1570-8705},
url = {https://doi.org/10.1016/j.adhoc.2022.102926},
doi = {10.1016/j.adhoc.2022.102926},
journal = {Ad Hoc Netw.},
month = sep,
numpages = {17}
}

@ARTICLE{nakayama2022resilience,
  author={Nakayama, Fernando and Lenz, Paulo and Nogueira, Michele},
  journal={IEEE Trans. on Netw. Serv. Manag.,}, 
  title={A Resilience Management Architecture for Communication on Portable Assisted Living}, 
  year={2022},
  volume={19},
  number={3},
  pages={2536-2548},
  doi={10.1109/TNSM.2022.3165729}}

@article{kashyap2025predictive,
author = {Kashyap, Vijaita and Ahuja, Rakesh and Kumar, Ashok},
title = {Predictive analysis-based load balancing and fault tolerance in fog computing environment},
year = {2025},
issue_date = {Apr 2025},
publisher = {Kluwer Academic Publishers},
address = {USA},
volume = {28},
number = {5},
issn = {1386-7857},
url = {https://doi.org/10.1007/s10586-024-04984-5},
doi = {10.1007/s10586-024-04984-5},
journal = {Cluster Computing},
month = apr,
numpages = {16}
}

@article{10.1145/3725985,
author = {Qi, Sheng and Feng, Haoyu and Liu, Xuanzhe and Jin, Xin},
title = {Efficient Fault Tolerance for Stateful Serverless Computing with Asymmetric Logging},
year = {2025},
issue_date = {May 2025},
publisher = {ACM},
address = {New York, NY, USA},
volume = {43},
number = {1–2},
issn = {0734-2071},
url = {https://doi.org/10.1145/3725985},
doi = {10.1145/3725985},
journal = {ACM Trans. Comput. Syst.},
month = jun,
articleno = {3},
numpages = {43}
}

@misc{wang2025ev,
      title={{EvoAgentX}: An Automated Framework for Evolving Agentic Workflows}, 
      author={Yingxu Wang and Siwei Liu and Jinyuan Fang and Zaiqiao Meng},
      year={2025},
      eprint={2507.03616},
      archivePrefix={arXiv},
      primaryClass={cs.AI},
      url={https://arxiv.org/abs/2507.03616}, 
}

@misc{wu2025git,
      title={Git Context Controller: Manage the Context of LLM-based Agents like Git}, 
      author={Junde Wu},
      year={2025},
      eprint={2508.00031},
      archivePrefix={arXiv},
      primaryClass={cs.SE},
      url={https://arxiv.org/abs/2508.00031}, 
}

@misc{sun2025agent,
      title={Towards Agentic Self-Learning {LLMs} in Search Environment}, 
      author={Wangtao Sun and Xiang Cheng and Jialin Fan and Yao Xu and Xing Yu and Shizhu He and Jun Zhao and Kang Liu},
      year={2025},
      eprint={2510.14253},
      archivePrefix={arXiv},
      primaryClass={cs.AI},
      url={https://arxiv.org/abs/2510.14253}, 
}

@article{Wu_2025,
title={{LLM}-Driven Agentic {AI} Approach to Enhanced {O-RAN} Resilience in Next-Generation Networks},
url={http://dx.doi.org/10.36227/techrxiv.174284755.59863143/v1},
DOI={10.36227/techrxiv.174284755.59863143/v1},
publisher={Institute of Electrical and Electronics Engineers (IEEE)},
author={Wu, Xingqi and Wang, Yuhui and Farooq, Junaid and Chen, Juntao},
year={2025},
month=mar }

@inproceedings{10.1145/3718958.3750505,
author = {Wang, Chenxu and others},
title = {Towards {LLM}-Based Failure Localization in Production-Scale Networks},
year = {2025},
isbn = {9798400715242},
publisher = {ACM},
address = {New York, NY, USA},
url = {https://doi.org/10.1145/3718958.3750505},
doi = {10.1145/3718958.3750505},
booktitle = {Proceedings of the ACM SIGCOMM 2025 Conference},
pages = {496–511},
numpages = {16},
location = {S\~{a}o Francisco Convent, Coimbra, Portugal},
series = {SIGCOMM '25}
}

@INPROCEEDINGS{10620727,
  author={Fang, Honglin and Zhang, Di and Tan, Can and Yu, Peng and Wang, Ying and Li, Wenjing},
  booktitle={IEEE INFOCOM 2024 - IEEE Conference on Computer Communications Workshops (INFOCOM WKSHPS)}, 
  title={Large Language Model Enhanced Autonomous Agents for Proactive Fault-Tolerant Edge Networks}, 
  year={2024},
  volume={},
  number={},
  pages={1-2},
  doi={10.1109/INFOCOMWKSHPS61880.2024.10620727}}

@ARTICLE{11169757,
  author={Qayyum, Adnan and Albaseer, Abdullatif and Qadir, Junaid and Al-Fuqaha, Ala and Abdallah, Mohamed},
  journal={IEEE Network}, 
  title={{LLM}-Driven Multi-Agent Architectures for Intelligent Self-Organizing Networks}, 
  year={2025},
  volume={},
  number={},
  pages={1-10},
  doi={10.1109/MNET.2025.3605319}}

@inproceedings{10.1145/3575879.3575976,
author = {Eleftherakis, George and Baxhaku, Fesal and Vasilescu, Anca},
title = {Bio-inspired Adaptive Architecture for Wireless Sensor Networks},
year = {2023},
isbn = {9781450398541},
publisher = {ACM},
address = {New York, NY, USA},
url = {https://doi.org/10.1145/3575879.3575976},
doi = {10.1145/3575879.3575976},
booktitle = {Proceedings of the 26th Pan-Hellenic Conference on Informatics},
pages = {116–122},
numpages = {7},
location = {Athens, Greece},
series = {PCI '22}
}

@inproceedings{10.1145/1315843.1315880,
author = {Balasubramaniam, Sasitharan and Botvich, Dmitri and Donnelly, William and Foghl\'{u}, M\'{\i}che\'{a}l \'{O} and Strassner, John},
title = {Biologically inspired self-governance and self-organisation for autonomic networks},
year = {2006},
isbn = {1424404630},
publisher = {ACM},
address = {New York, NY, USA},
url = {https://doi.org/10.1145/1315843.1315880},
doi = {10.1145/1315843.1315880},
booktitle = {Proceedings of the 1st International Conference on Bio Inspired Models of Network, Information and Computing Systems},
pages = {30–es},
location = {Cavalese, Italy},
series = {BIONETICS '06}
}

@inproceedings{10.1145/1570256.1570291,
author = {Samie, Mohammad and Dragffy, Gabriel and Pipe, Tony},
title = {Novel bio-inspired self-repair algorithm for evolvable fault tolerant hardware systems},
year = {2009},
isbn = {9781605585055},
publisher = {ACM},
address = {New York, NY, USA},
url = {https://doi.org/10.1145/1570256.1570291},
doi = {10.1145/1570256.1570291},
booktitle = {Proceedings of the 11th Annual Conference Companion on Genetic and Evolutionary Computation Conference: Late Breaking Papers},
pages = {2143–2148},
numpages = {6},
location = {Montreal, Qu\'{e}bec, Canada},
series = {GECCO '09}
}

@inproceedings{10.1145/1388969.1389045,
author = {Madureira, Ana and Santos, Filipe and Pereira, Ivo},
title = {Self-managing agents for dynamic scheduling in manufacturing},
year = {2008},
isbn = {9781605581316},
publisher = {ACM},
address = {New York, NY, USA},
url = {https://doi.org/10.1145/1388969.1389045},
doi = {10.1145/1388969.1389045},
booktitle = {Proceedings of the 10th Annual Conference Companion on Genetic and Evolutionary Computation},
pages = {2187–2192},
numpages = {6},
location = {Atlanta, GA, USA},
series = {GECCO '08}
}

@article{10.1145/1152934.1152937,
author = {Babaoglu, Ozalp and others},
title = {Design patterns from biology for distributed computing},
year = {2006},
issue_date = {September 2006},
publisher = {ACM},
address = {New York, NY, USA},
volume = {1},
number = {1},
issn = {1556-4665},
url = {https://doi.org/10.1145/1152934.1152937},
doi = {10.1145/1152934.1152937},
journal = {ACM Trans. Auton. Adapt. Syst.},
month = sep,
pages = {26–66},
numpages = {41}
}

@article{10.1145/2168260.2168276,
author = {Fisch, Dominik and Fisch, Dominik and J\"{a}nicke, Martin and Kalkowski, Edgar and Sick, Bernhard},
title = {Techniques for knowledge acquisition in dynamically changing environments},
year = {2012},
issue_date = {April 2012},
publisher = {ACM},
address = {New York, NY, USA},
volume = {7},
number = {1},
issn = {1556-4665},
url = {https://doi.org/10.1145/2168260.2168276},
doi = {10.1145/2168260.2168276},
journal = {ACM Trans. Auton. Adapt. Syst.},
month = may,
articleno = {16},
numpages = {25}
}

@INPROCEEDINGS{11064129,
  author={K, Thinakaran and Kanagasabapathi, K. and A, Vishnuprasath. and R, Rajagopal and Tripathi, Aanjey Mani and G, Guna},
  booktitle={2025 3rd International Conference on Communication, Security, and Artificial Intelligence (ICCSAI)}, 
  title={Developing Self-Healing Networks with Bio-Inspired Algorithms: Enhancing Resilience in Modern Networks}, 
  year={2025},
  volume={3},
  number={},
  pages={865-871},
  doi={10.1109/ICCSAI64074.2025.11064129}}

@INPROCEEDINGS{11196711,
  author={Husain, Saef Obad and Narayana, V. A. and Annapurna, Thammishati and Kumar, B. Sampath and Shunmugapriya, V. and Srihari, T.},
  booktitle={2025 International Conference on Metaverse and Current Trends in Computing (ICMCTC)}, 
  title={A Bio-Inspired Self-Healing Machine Learning Framework for Autonomous Fault Recovery in Computational Networks}, 
  year={2025},
  volume={},
  number={},
  pages={1-4},
  doi={10.1109/ICMCTC62214.2025.11196711}}

@misc{donta2025resilientdesignactive,
      title={Resilient by Design -- Active Inference for Distributed Continuum Intelligence}, 
      author={Praveen Kumar Donta and Alfreds Lapkovskis and Enzo Mingozzi and Schahram Dustdar},
      year={2025},
      eprint={2511.07202},
      archivePrefix={arXiv},
      primaryClass={cs.DC},
      url={https://arxiv.org/abs/2511.07202}, 
}

@data{8wf2-2y40-24,
doi = {10.21227/8wf2-2y40},
url = {https://dx.doi.org/10.21227/8wf2-2y40},
author = {Nutt Chairatana and Rathachai Chawuthai},
publisher = {IEEE Dataport},
title = {Cloud Stateless System Performance Metrics and Status},
year = {2024} }

@misc{models,
author = {OpenAI},
title = {{OpenAI} models},
year ={2025},
 howpublished = {\url{https://platform.openai.com/docs/models}},
note = {Last accessed: \today}
}

@misc{pydantic,
author = {Pydantic},
title = {pydantic},
year ={2025},
 howpublished = {\url{https://docs.pydantic.dev/latest/api/base_model/}},
note = {Last accessed: \today}
}

@misc{langchain,
author = {LangChain},
title = {LangChain},
year ={2025},
 howpublished = {\url{https://www.langchain.com/}},
note = {Last accessed: \today}
}

@misc{sts,
author = {huggingface},
title = {all-MiniLM-L6-v2},
year ={2025},
 howpublished = {\url{https://huggingface.co/sentence-transformers/all-MiniLM-L6-v2}},
note = {Last accessed: \today}
}

@INPROCEEDINGS {10301257,
author = { Zhu, Jieming and He, Shilin and He, Pinjia and Liu, Jinyang and Lyu, Michael R. },
booktitle = { 2023 IEEE 34th International Symposium on Software Reliability Engineering (ISSRE) },
title = {{ Loghub: A Large Collection of System Log Datasets for AI-driven Log Analytics }},
year = {2023},
volume = {},
ISSN = {},
pages = {355-366},
doi = {10.1109/ISSRE59848.2023.00071},
url = {https://doi.ieeecomputersociety.org/10.1109/ISSRE59848.2023.00071},
publisher = {IEEE Computer Society},
address = {Los Alamitos, CA, USA},
month =Oct}

@INPROCEEDINGS{11270701,
  author={Donta, Praveen Kumar and Zhang, Qiyang and Dustdar, Schahram},
  booktitle={2025 IEEE 11th World Forum on Internet of Things (WF-IoT)}, 
  title={Performance Measurements in the {AI}-Centric Computing Continuum Systems}, 
  year={2025},
  volume={},
  number={},
  pages={1-6},
  location={Chengdu, China},
  doi={10.1109/WF-IoT64238.2025.11270701}}

@inproceedings{10.1145/3650212.3652123,
author = {Jiang, Zhihan and Liu, Jinyang and Huang, Junjie and Li, Yichen and Huo, Yintong and Gu, Jiazhen and Chen, Zhuangbin and Zhu, Jieming and Lyu, Michael R.},
title = {A Large-Scale Evaluation for Log Parsing Techniques: How Far Are We?},
year = {2024},
isbn = {9798400706127},
publisher = {ACM},
address = {New York, NY, USA},
url = {https://doi.org/10.1145/3650212.3652123},
doi = {10.1145/3650212.3652123},
booktitle = {Proceedings of the 33rd ACM SIGSOFT International Symposium on Software Testing and Analysis},
pages = {223–234},
numpages = {12},
location = {Vienna, Austria},
series = {ISSTA 2024}
}
